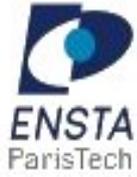

École Nationale Supérieure de Techniques Avancées

*Master Thesis Report*

# Human Posture Recognition and Gesture Imitation with a Humanoid Robot

*By*

## Amir ALY

**(Master Student- Université Pierre et Marie Curie- UPMC)**

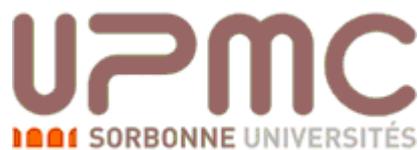

*Supervisor*

**Adriana TAPUS**

**(Cognitive Robotics Laboratory – ENSTA Paris-Tech)**

# _ACKNOWLEDGMENT_

In the first place, I would like to express my sincere appreciation to **_Prof. Adriana TAPUS_**, cognitive robotics laboratory, **_ENSTA Paris-Tech_**, for her kind supervision, generous advice, and support during the whole scope of this work.

Also, it is my pleasure to pay tribute to all the members of the cognitive robotics laboratory in **_ENSTA Paris-Tech_**, who provided me unflinching encouragement and support in various ways that enabled me to finish the mentioned target of my research successfully.

Similarly, i am very grateful to all the technical team members of **_ENSTA Paris-Tech,_** for the helpful technical facilities and cooperation they showed during my research.

Last but not least, many thanks to all my professors in Pierre and Marie Curie University, **_UPMC_**, for the attention, care and cooperation i found during my Master study in this reputable university.

# ABSTRACT

Autism is a highly variable neurodevelopmental disorder characterized by impaired social interaction and communication , and by restricted and repetitive behavior. The problematic point concerning this neurodevelopmental disorder is its causes which are unknown till now, and therefore it cannot be treated medically.

Recently, robots have been involved in the development of the social behavior of autistic children who showed a better interaction with robots than with their peers. One of the striking social impairments that is widely described in autism literature is the deficit of imitating the others [Rogers and Pennington 1991; Williams et al., 2004]. Trying to make use of this point, therapists and robotic researchers have been interested in designing triadic interactional *(Human – Robot- Child)* imitation games, in which the therapists starts to perform a gesture then the robot imitates it, and then the child tries to do the same, hoping that these games will encourage the autistic child to repeat these new gestures in his daily social life.

The robot used in the imitational scenarios has to be able to well understand and classify different gestures and body postures using its own sensors (in our case, its camera).

In this work, the control of the robot is performed by an external computer communicating with the robot via a *TCP/IP* socket, through which the filmed gesture by the robot's camera will be transferred to the computer to be analyzed and classified by *MATLAB*, which in turn controls the robot to imitate a specific pre-programmed gesture correspondent to the number of the class attributed to the tested gesture.

The robot used in this research is *NAO (developed by ALDEBARAN robotics)* .It has articulations and degrees of freedom similar to those of a human, and all the gestures and body postures to imitate are programmed by python language.

This report is composed of 3 chapters that indicate in details all the previous discussed points, and of some indicative appendixes that can be found at the end of the report.

# LIST OF CONTENTS

# LIST OF FIGURES

# LIST OF TABLES

# Chapter 1

## Static gestures recognition

The static gestures studied in this chapter cast light mainly on hand gestures, and how to extract the hand from a binary image in order to calculate the characterization vector necessary for the classification. The word "static" means that the gesture will be analyzed from a picture, not from a video like in the dynamic gestures *(see chapter 2)*, which makes it, kindly, more easy.

The database of static gestures is composed of 7 pictures of hand gestures taken for 8 persons. These gestures are illustrated in the figure below where the extracted contours of the hand making different gestures are illustrated.

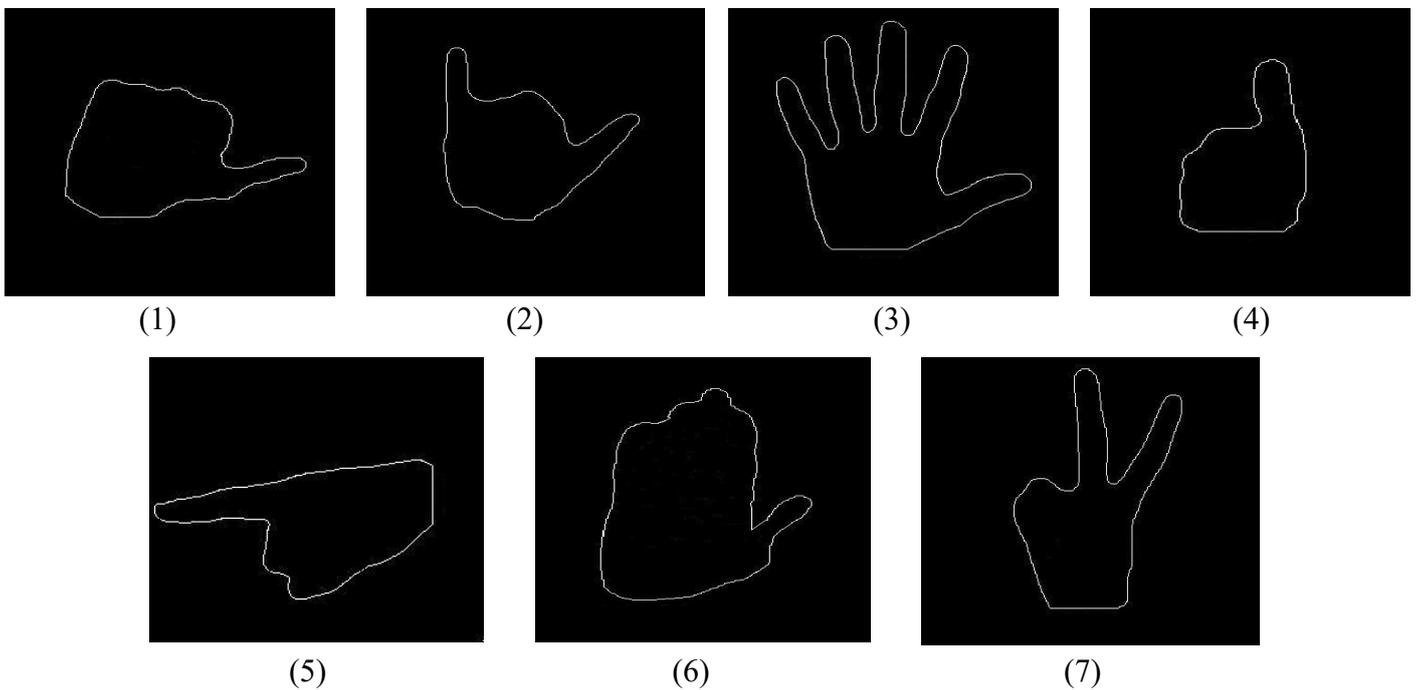

**Fig. (1-1)** Seven static gestures

The total efforts of this work mainly focused on analyzing video captured gestures *(see chapter 2)* in the context of making the robot able to imitate dynamic gestures. However, static gestures also occupied an important part our *HRI* application because these static gestures can convey specific communicating messages like: perfect gesture, pointing gesture, and stop gesture etc. "gestures *4 & 5 & 6 respectively*", which in turn obligate *HRI* researchers in the world to cast more light on them for their importance.

The proposed method's steps used to analyze and classify static gestures could be summarized in the following outline:

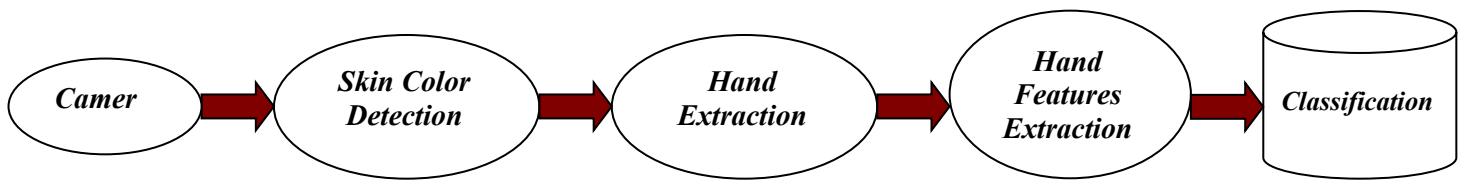

**Fig. (1-2)** Static gestures recognition outline

# (1) Skin color detection

Skin-color detection has been employed in many applications such as face detection, gesture recognition, human tracking etc. It has been proved that using the skin-color information in the pre-process can reduce the difficulties of problems. But, due to the effects of different human races, ambient lights and confusing backgrounds, detecting skin-color accurately is not an easy task [4] [18].

In order to develop a successful skin detector model, it is necessary to transfer the image *RGB* color model to *YCbCr* color space *(Y presents the Luma component and Cb & Cr are the blue and red chroma components)* which consumes less space while retaining the full perceptual value, consequently, processing an image in the *YCbCr* color space will be much more faster and efficient than in the *RGB* color space.

When the color model is presented in the *YCbCr* space, the values of the blue and red chroma could be real indicators for skin regions regardless of the race of the skin *(Meanwhile, Luma component is not important in this issue because it highly varies from one person to another according to light effects)*. These values could be summarized in the following intervals:

$$77 \leq C_b \leq 127 \quad \& \quad 133 \leq C_r \leq 173 \qquad (1)$$

Despite the relevance and pertinence of the previous intervals in characterizing skin regions regardless of the race, there may appear some special difficulties which could be summarized in the following points:

## (1-1) Effect of the Background on the detection of skin regions

The background may intervene in the detected skin regions if its color is near to the color of the skin, so that it gets inside the range of the skin detecting intervals mentioned above, and be considered as a part of the skin regions. The following figure reveals that effect [9].

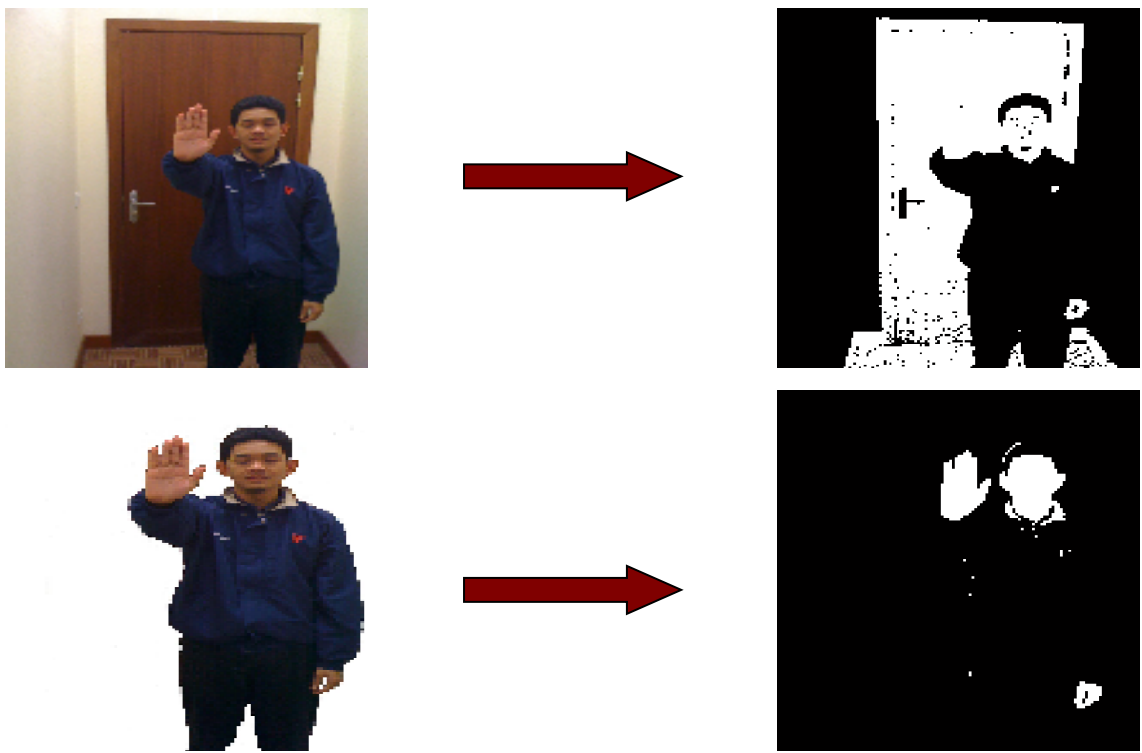

**Fig. (1-3)** Background effect on the detection of skin regions

The above figure indicates two colors for the background; the first photo has a background color near to the color of the hand and face of this colleague and so, it gets considered as skin as indicated in the corresponding binary image. While in the second photo, the original background was removed and the color of the photo's background is completely white, so skin regions are perfectly detected in this case as indicated in the binary image without any negative effect from the background. Therefore, the easiest solution for this problem is to use special backgrounds of colors highly contrasted to the possible colors of skin like: *pure red, pure bleu, pure green, pure white or pure black*, as indicated in the previous figure.

### (1-2) Effect of light on the detection of skin regions

Light also may affect negatively the detected skin regions and consequently, the extracted hand contour at the end which affects badly, in turn, the classification process. The following figure indicates this effect.

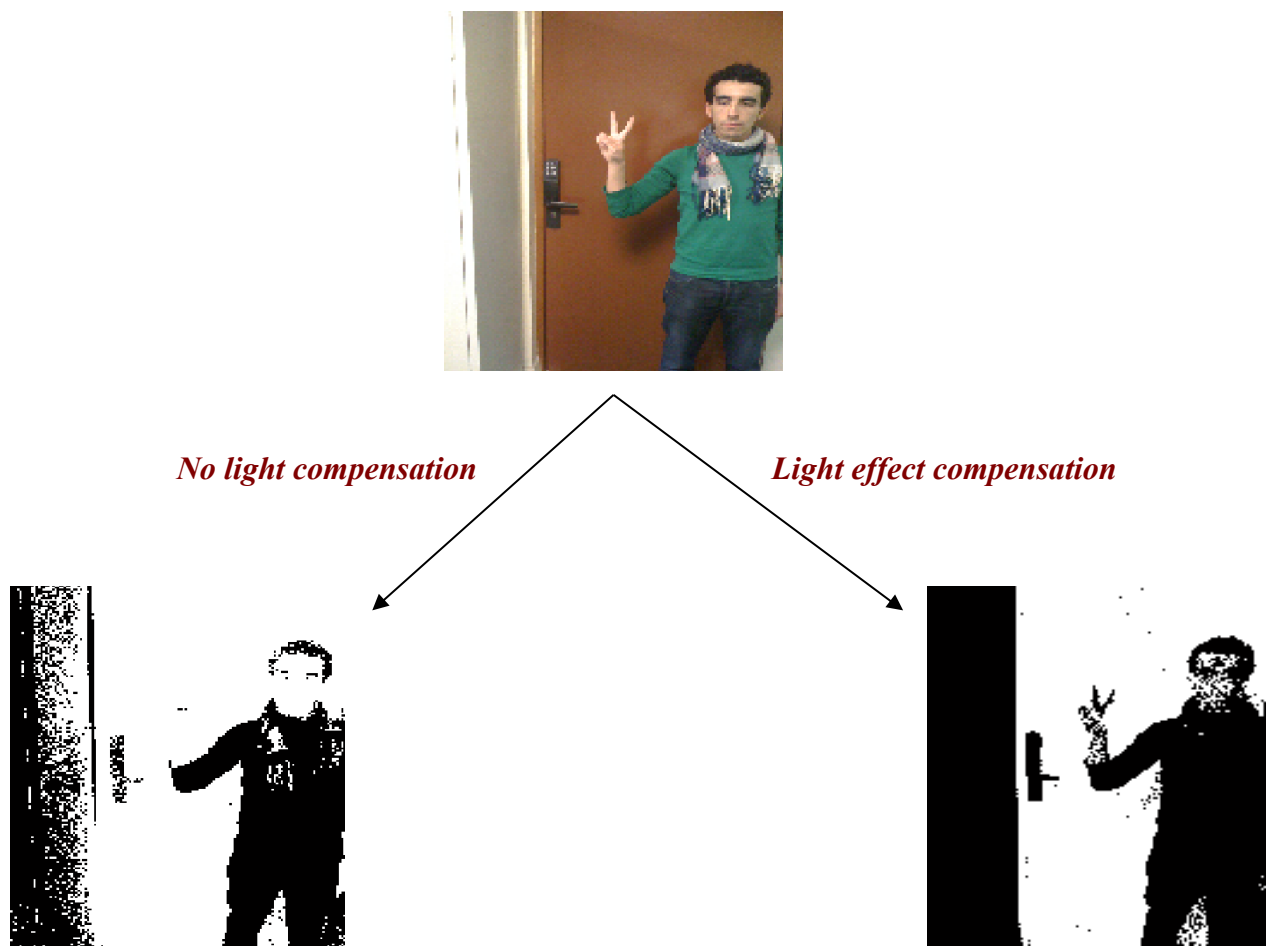

**Fig. (1-4)** Light effect on the detection of skin regions

The above figure discusses the effect of light compensation on detecting skin regions. In fact, the uncovered parts of the body *(face and hand)* still partially confused with the background color, which confirms the high necessity of using a highly contrasted background color as mentioned before.

On the other hand, the effect of light compensation is so clear in the two binary images in the above figure. The algorithm used in compensating light effects is called *"Gray World"* which is applied to the image before transferring its color space from RGB to *YCbCr*.

This algorithm would be summarized in the following steps:

1- Calculate the inverse of the average values of the *R, G, and B* color components.
2- Calculate the maximum value of the inverse averages calculated in *step 1*.
3- Calculate the scaling factors by dividing *steps 1* over *2* for the three color components *R, G, and B*.
4- Multiply each color component *R, G, and B* by the corresponding scaling factor calculated in *step 3*.

Despite the importance of compensating light effects, but in some situations like when we have a highly contrasted color background, we will find that there is no big difference between compensating and non compensating light effects, because the highly contrasted background will reduce the bad effects of light *on the background and on the person himself*, and as a result the detected skin regions will be of reasonable perfection. In all cases, integrating light compensation to skin detection regardless of the color of the background will not be, at least, harmful.

The following figure corresponds to same person in shown figure (4) after removing the background. The two binary images reveal that there is no real difference *(just in this case)* after and before light compensation.

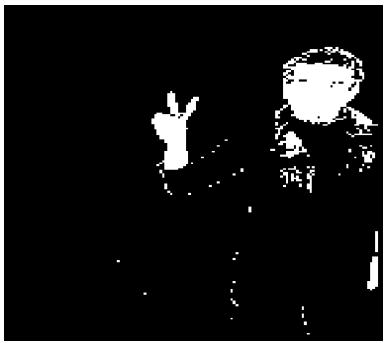
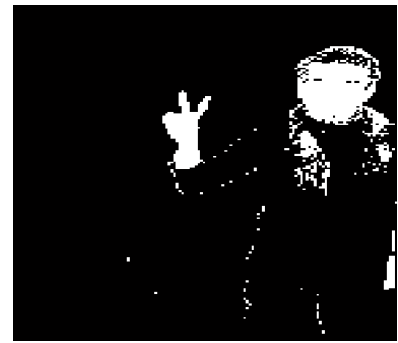

*Before light compensation*            *After light compensation*

**Fig. (1-5)** Utility of light compensation in the existence of a contrasted background

## (1-3) Effect of clothes color on the detection of skin regions

Clothes color also has to be contrasted to the color of skin, *if possible*, to avoid all possible wrong skin detected regions as indicated in the last figure where the scarf was considered wrongly as a part of the detected skin regions.

So, the required steps for successfully detecting skin region could be summarized as:

- Highly contrasted background color to the color of skin.
- Light compensation.
- Clothes of contrasted color to the color of skin.

The above difficulties when detecting skin regions had oriented researchers' minds towards the necessity to find out other robust criteria to characterize skin, which lead to the *"adaptive detection of skin regions"*. However, it hasn't approved, for the moment, its complete efficiency in processing complex color backgrounds. Therefore, it seems indispensable to use backgrounds of specific colors highly contrasted to human skin color *(as listed in 1-1)* in order to precisely detect skin regions.

# (2) Hand extraction

After detecting skin regions, it is necessary to extract the hand from the captured image in order to start characterizing its contour, which is necessary for the classification. Before this step, some filtration has to be done to the image for the noise that may appear [11].

## (2-1) Noise filtration

Two types of noise may appear in the detected skin regions image according to the complexity of the image conditions, like: light effect, clothes' colors, and background color and homogeneity, these types of noise are:

1. Background pixels considered as skin pixels.
2. Pixels inside the hands or the face *(the uncovered regions of the body)* being erroneously identified as non-skin pixels.

Noise of type 1, is considered the most famous and frequent type of noise affecting skin detected regions. It varies from an image to another according to the color and complexity of the background as mentioned before. The first step in minimizing this effect is to calculate the connected components in the binary image *(where the detected skin pixels are in white color and the rest background pixels are in black color)*, then calculate the size of each component in the image. Normally, if the surrounding environment is in ideal conditions *(suitable background color and less light effect)* , the size of noise regions *(which are mostly unconnected to the uncovered parts of the body)* will be small enough to be eliminated by a suitable threshold value compared to the size of the other detected skin regions like hands and face. To the contrary, if the negative effects of light or background color exist, the size of noise pixels will be bigger and, consequently, these regions will be connected to the hands or the face regions which makes the process of eliminating the noise more difficult , like using morphological operations or etc.

Definitely, fixing a threshold size value or deciding which morphological operation to use depends on studying first the environment and all relevant conditions then taking some testing photos for many persons and analyzing them to arrive to an idea about the necessary processing to do.

The following figure indicates a photo taken over an open background in the absence of light effect. It is clear that the small unconnected noise populations are easy to be eliminated by applying a suitable threshold value.

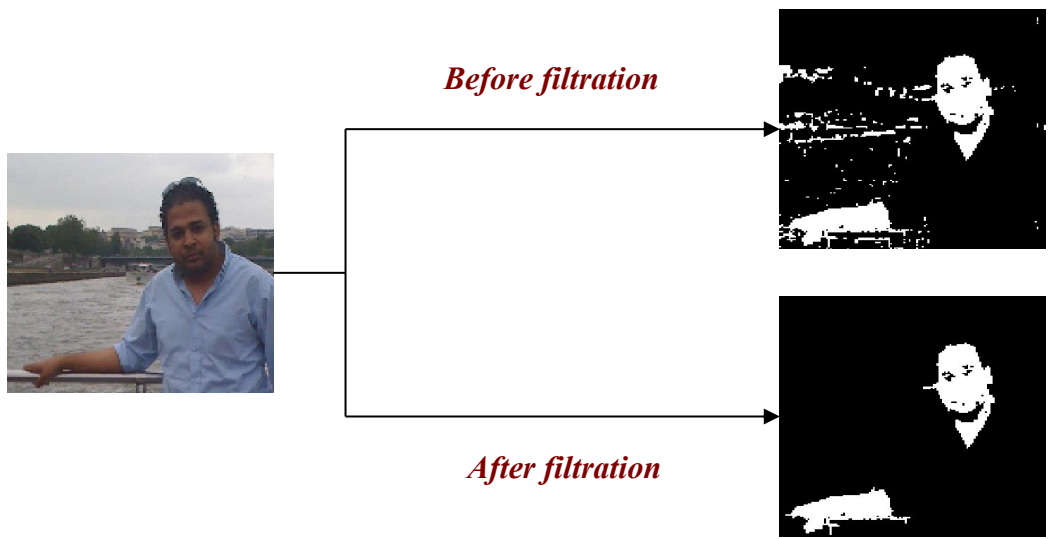

**Fig. (1-6)** Elimination of small noise populations

While in the following figure it is clear that noise pixels are cumulated in bigger regions connected to the head and the hand of Bill Gates, consequently, thresholded filtration of these regions will not be useful and applying morphological operation will be necessary.

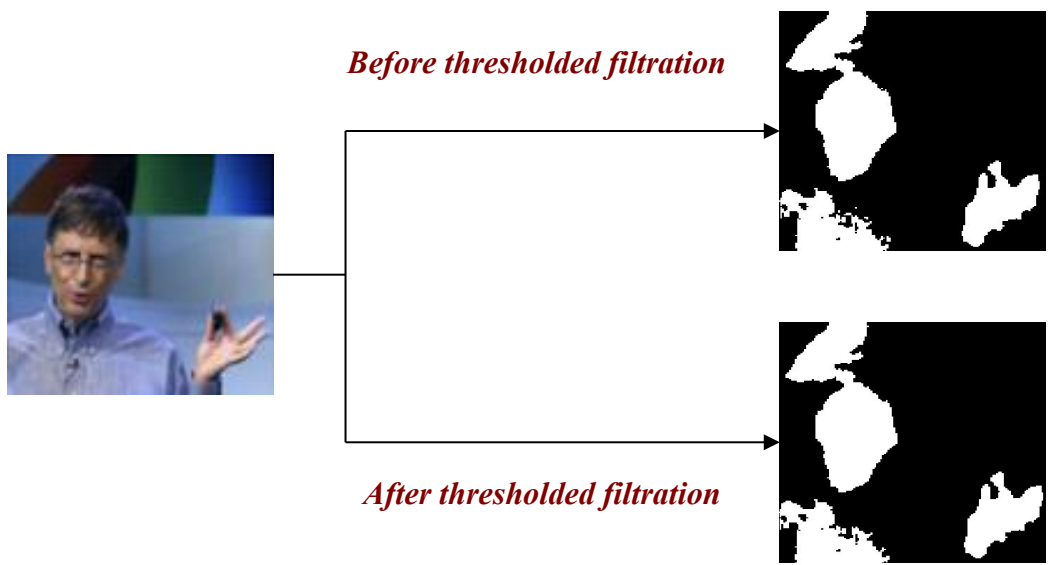

**Fig. (1-7)** Bigger noise regions need more complicated processing

The previous figures are just examples for different situations that may exist while trying to filter the image from undesired details and noise that may hinder the extraction of hand contour, and as mentioned before, these undesired details vary from an image to another in terms of the background color and light effect.

Noise of type 2 is less frequent with respect to the first noise type. It involves some pixels considered as non skin pixels inside the face *(like the mouth, eyes, and eyebrows)* or the hand. Logically, pixels of this noise type are of small sizes and could be corrected easily by suitable thresholding; i.e. pixels' populations of sizes inferior to a certain threshold value will be considered as type 2 noise pixels and will be corrected by changing their value from 0 *"black"* to 1 *"white"* as indicated in the following figure which casts light on the undesired details in the face which are disappeared after correction.

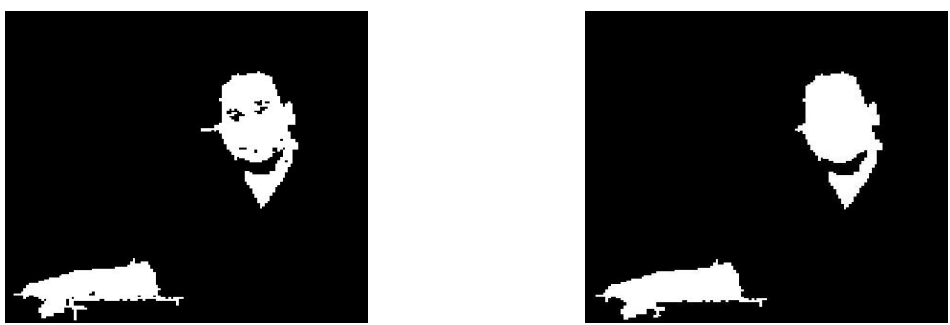

**Fig. (1-8)** Correction of skin pixels considered as non skin pixels

### (2-2) Distance transformation

It is a derived representation of a digital image in which it labels each pixel of the image with the distance to the nearest obstacle pixel. Many distance metrics could be used in this calculation, but in this work, the Chamfer distance will be the distance metric under study. The following figure illustrates the Chamfer distance which is a two pass distance algorithm.

Chamfer distance transformation image *DT* will be calculated by detecting the pixels' coordinates of the binary image components *(hands and head)* and giving them label name =0 as being foreground pixels, then the distance metrics indicated in the following figure, will be applied to all the surrounding pixels of the up mentioned image components in two scanning directions; forward and backward as indicated below [11] [5].

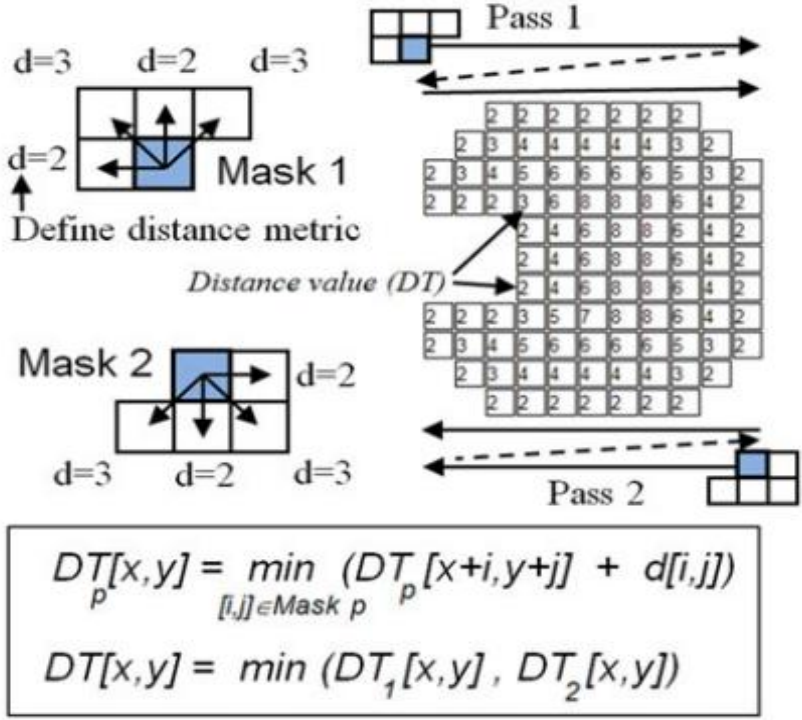

**Fig. (1-9)** Chamfer distance metric

The main target of this distance transformation is to extract some pixels called *"distance based feature pixels"* which help in distinguishing the hand from the other components in the binary image and consequently, facilitating its extraction. These pixels could be identified easily in the distance transformation image as the pixels that satisfy the following equation [11]:

$$\sum CMP(DT[x,y] \geq DT[x+i, y+j]) = 8 \quad \text{With} \quad 1 \geq [i,j] \geq -1 \quad (2)$$

Where *i, j* are integers, and $DT[x,y]$ and $DT[x+i, y+j]$ are correspondingly the distance values at the position $[x,y]$ and $[x+i, y+j]$ in the distance transformation image. The comparison function *CMP* defines that $CMP(DT[x,y] \geq DT[m,n]) = 1$ if the comparison expression is true and 0 otherwise.

Figure (10) illustrates the detection of the distance based feature pixels. It is clear that each feature pixel could be considered as a local maximum in a surrounding of 8 pixels. Feature pixels of biggest values are mostly located on the middle of the hand, and the lowest values feature pixels are mostly located at the beginning of the wrist, while feature pixels of fingers have always intermediate values, as indicated in the figure below.

Obviously, some feature pixels may appear on the face or even on some noise regions in the image, but the number of the feature pixels located on the hand *(especially if the hand is open)* is usually much bigger than the number of similar pixels located on the face or other regions, which facilitates, in turn, distinguishing and extracting only the hand from the image.

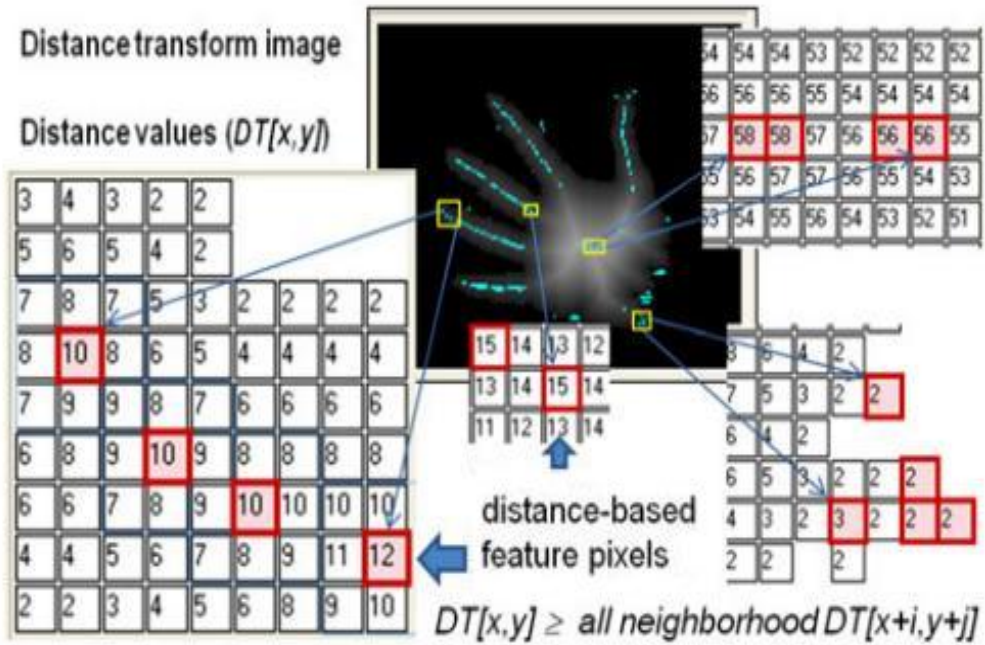

**Fig. (1-10)** Possible values of the distance based feature pixels located on the hand

The previous figure illustrates an example for the possible values of the feature pixels located in the hand, these values depends on the size of the image and the distance between the camera and the hand making gesture. In this work, i choosed only the distance based feature pixels whose values are limited to this interval [11] [10]:

$$4 \leq \text{Pixel's value} \leq 12 \qquad (3)$$

The previous interval will mostly eliminate some feature pixels on the hand *(at the middle and at the beginning of the wrist)* beside other unnecessary pixels on the face. However, there may stay some feature pixels around the hand not eliminated, so, it is necessary to extract only the hand whatever being alone or attached to a part of the arm, to eliminate these unnecessary pixels around the hand. In order to do that, it is necessary to calculate the centroid *(refer to equation 8)* of the hand, then extract it and reposition it at the middle of a new frame *(see figure 11)*.

In case of some complex backgrounds where there are many persons, it is possible to find that the hand making gesture to extract connected to another object *(hand or head)* of another person in the distance image. Therefore, some processes have to be done like:

- Apply a suitable morphological operation *(Erosion)* with a suitable threshold value to separate the connected objects. This threshold will be detected according to the complexity of the background.

- Count the number of feature pixels in each object in the eroded image, and basing on the fact that *the open hand with stretched fingers mostly has the biggest number of feature pixels than the other part of the body like closed hands or head*, it will be easy to specify a feature pixels number as a threshold value to distinguish between objects in the image and to extract only the hand and position it in a new image frame. This threshold value is precised experimentally according to the conditions of the processed image.

The following figure shades on the detected feature pixels on the hand and a small part of the arm, and shows the hand after being extracted and positioned at the middle of a new frame.

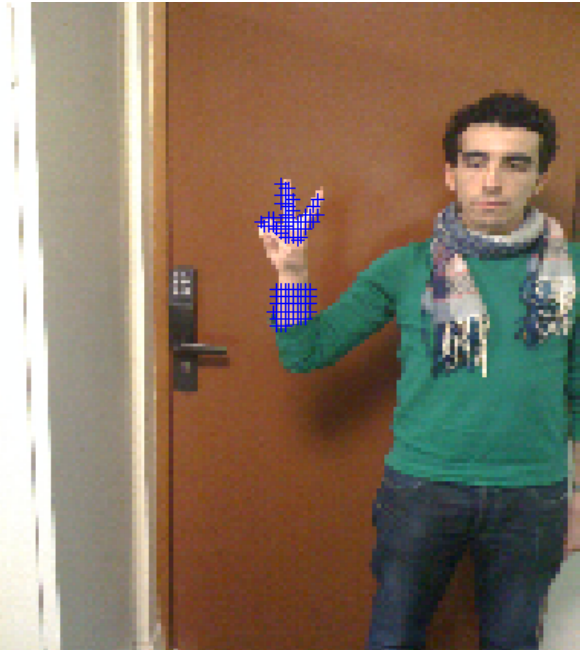 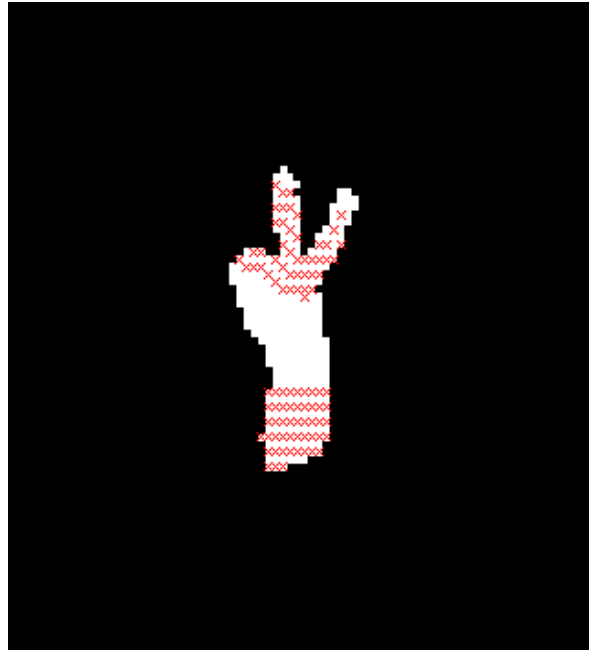

**Fig. (1-11)** Distance based feature pixels of the hand

In the figure above, it is clear that there are some feature pixels located on the part of the arm attached to the hand making gesture. These unnecessary pixels will be eliminated by eliminating that small part of the arm and retaining only the hand with the feature pixels located on fingers, and this will be discussed in the following part of this chapter.

**(2-3) Retaining the hand and distance based feature pixels on fingers**

The following figure illustrates the extraction of the hand with its feature pixels and eliminating all the other unnecessary parts of the image. Three main points have to be calculated [11] [10]:

1. *The centroid*: which could be calculated as explained in equation (8) in this chapter.

2. *The upper feature pixels' mean point*: In order to calculate this point, it is necessary to discard all feature pixels located out of the hand, which is the main target to extract, like these pixels located on the small part of the arm connected to the hand. Therefore, i used the coordinates of the centroid and discarded all the feature pixels that have vertical coordinates bigger than the vertical coordinates of the centroid, then i calculated the mean point of the retained feature pixels located on the fingers as indicated in the following figure.

3. *The point P[Xp, Yp]* : at which the separating bleu line is drawn, is calculated in terms of the centroid and mean points as following:

$$X_p = X_c - round(0.75*(X_m - X_c)) \quad (4)$$

$$Y_p = Y_c - round(0.75*(Y_m - Y_c)) \quad (5)$$

The coming figure illustrates the extraction of the hand and the elimination of the unnecessary down part. The red line which passes through the calculated three points presents the direction of the hand, while the blue line is the perpendicular to the red line passing though point p, and presents the separating line between the retained part and the eliminated part.

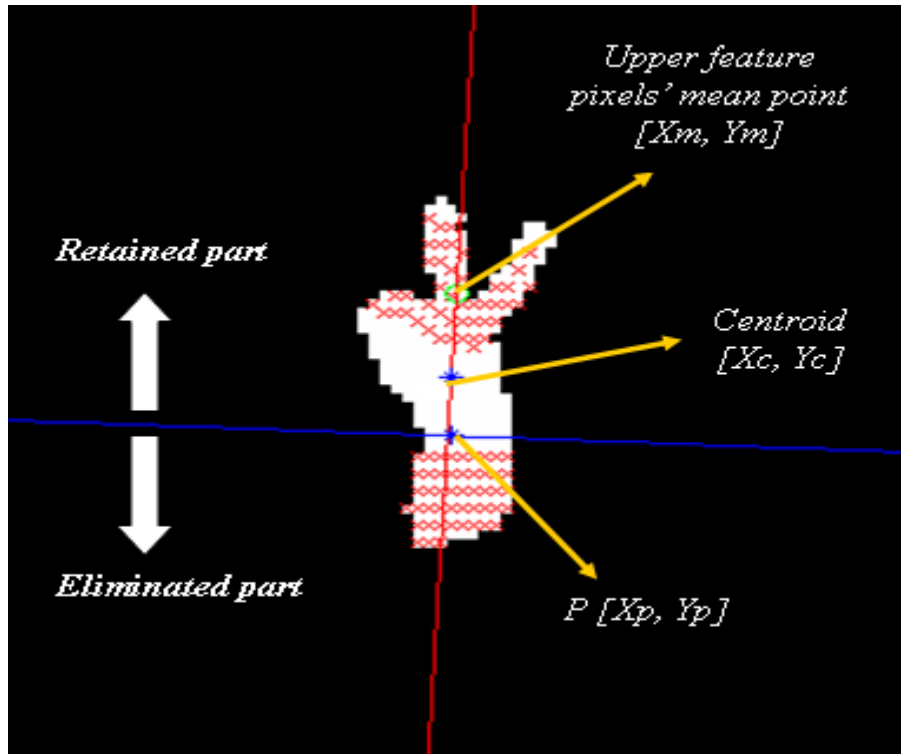

**Fig. (1-12)** Extracting the hand and eliminating unnecessary parts

The following figure reveals the hand after being extracted. When eliminating the part of arm connected to the hand, it is necessary to recalculate the centroid of the final extracted hand and reposition the hand again at the middle of the image frame as indicated below.

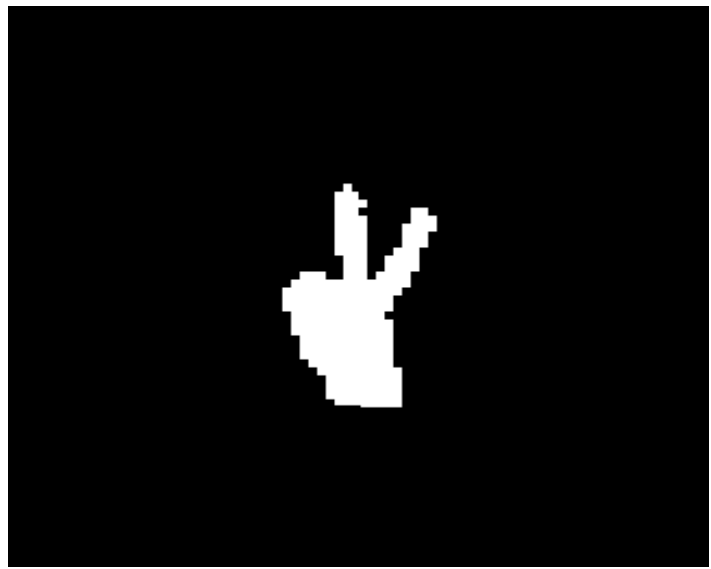

**Fig. (1-13)** Extracted hand

The last remaining step is to extract the outer contour of the hand in order to characterise it as explained in the next point in this chapter. I used *Sobel* algorithm for edge detection to detect the contour of the hand by calculating the gradient of image pixels, and choosing the pixels of maximum gradient which are the pixels of the contour, as indicated in the following figure.

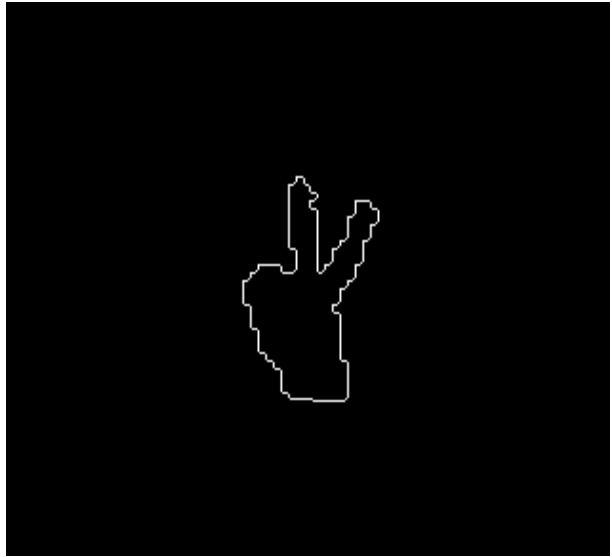

**Fig. (1-14)** Extracted hand's contour

The previous figure reveals the extracted contour of the hand. It is clear that the contour needs a refining process; otherwise the classification process may be affected badly because it depends on characterizing the contour geometrically and so, this zigzag line may cause a negative effect [18].

In this work, the camera used to photo the gestures had a high resolution of *1600 \*1200* pixels, so, to process an image of that resolution it will take some time. Therefore, i made some qualitative minimization for the resolution of the image to be *176\*144* pixels; i.e. smaller dimensions with a near quality, using the cubic interpolation. As long as the source image that been processed was interpolated, it is logical to find the edges of the hand not very smooth *(i.e. in a zigzag line)* compared to the original image before interpolation *(see figure 14)*.

Therefore, it is necessary to make another smoothing process for the previous extracted contour using curve fitting, in which a small window will move along the contour and will replace its pixels with other fitted pixels as indicated in the figure below. This smoothed contour will be characterized geometrically as indicated in the following part of this chapter.

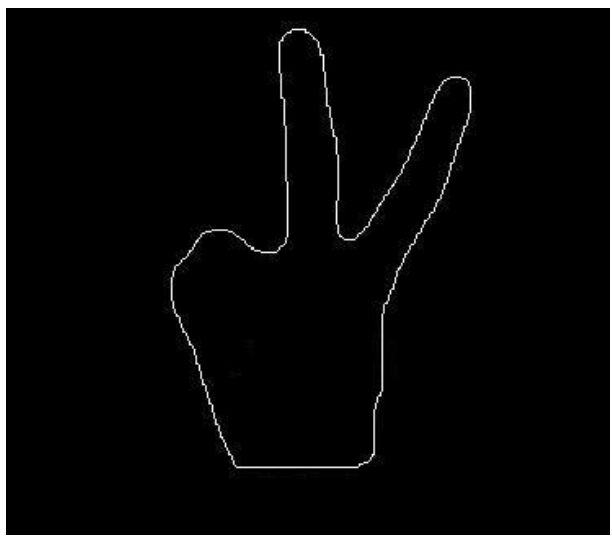

**Fig. (1-15)** Extracted hand's smoothed contour

## (3) Hand features extraction

After extracting the hand from the binary image, it is necessary to characterise it in a pertinent way in order to start classifying gestures. In general, in order to characterise a hand making gesture, two main approaches could be used [16]:

- *3D hand model based method*: which studies parameters like; joint angles and palm position.
- *Appearance based method*: which studies geometrical parameters and fingertip position.

In this work, i used the appearance based method to characterise the hand in terms of its contour geometrical parameters like: area, perimeter, and geometrical moments, as listed below in the characteristic vector [18]:

$$Characteristic\ vector = [\ \frac{P^4}{I_{min}}\ ,\frac{A^2}{I_{max}},\ \frac{A^2}{I_{min}}\ ,\frac{D_{max}}{D_{min}},\ \frac{P^2}{A}\ ,\ \frac{I_{min}+I_{max}}{A^2}\ ,\ \frac{I_{max}-I_{min}}{I_{max}+I_{min}}\ ] \qquad (6)$$

### (3-1) Calculation of the characteristic vector's parameters

1. The area and perimeter parameters are easy to calculate by *MATLAB* using instructions *bwarea*, and *bwperim* in turn.
2. $I_{max}$, $I_{min}$ are the maximum and minimum principal second moments, and could be calculated as following:

- The central moments of order $(p+q)$ for an image $f(x,y)$ could be defined as:

$$\mu_{pq} = \iint_{-\infty,\infty} (x-\bar{x})^p (y-\bar{y})^q f(x,y)dxdy \quad ,\text{Where } \bar{x},\ \bar{y} \text{ are the coordinates of the centroid} \qquad (7)$$

- The centroid coordinates $\bar{x}$ and $\bar{y}$ could be calculated in terms of the raw moments as following:

$$\bar{x} = \frac{M_{10}}{M_{00}} \quad \text{and} \quad \bar{y} = \frac{M_{01}}{M_{00}} \qquad (8)$$

- The raw moments $M_{pq}$ could be calculated by:

$$M_{pq} = \iint_{-\infty,\infty} (x)^p (y)^q f(x,y)dxdy \qquad (9)$$

- Last but not least, $I_{max}$ and $I_{min}$ could be calculated by:

$$I_{max} = \frac{\left(\mu_{20}+\mu_{02}+sqrt\left(4\mu_{11}^2 + (\mu_{20}-\mu_{02})^2\right)\right)}{2} \qquad (10)$$

$$I_{min} = \frac{\left(\mu_{20}+\mu_{02}-sqrt\left(4\mu_{11}^2 + (\mu_{20}-\mu_{02})^2\right)\right)}{2} \qquad (11)$$

3- $D_{max}, D_{min}$ are the maximum and minimum distance from the centre of gravity, and could be calculated as following:

- In the binary image, find the coordination of the foreground pixels *(i.e. pixels of values =1)*.

- Calculate the absolute difference between the x and y coordinates of the foreground pixels and the corresponding x and y coordinates of the centroid, then precise the maximum and minimum values in each direction *(x_diff_max, y_diff_max and x_diff_min, y_diff_min)*.

- Substitute with the previous calculated values in the following equations:

$$D_{max} = sqrt(x\_diff\_max^2 + y\_diff\_max^2) \quad (12)$$

$$D_{min} = sqrt(x\_diff\_min^2 + y\_diff\_min^2) \quad (13)$$

4- The parameters, $\dfrac{p^2}{A}$, $\dfrac{I_{min}+I_{max}}{A^2}$, and $\dfrac{I_{max}-I_{min}}{I_{max}+I_{min}}$ are called, in turn, the compactness, spreadness, and elongation of the hand contour, which make, *as clear in their meanings*, the characteristic vector more relevant and better expressing the hand contour.

## (4) Classification

After constructing the database using the characteristic vector of each of the seven gestures mentioned at the beginning of this chapter *(see figure 1)*, the k- nearest neighbour classification method *(KNN)* supported by the cross validation were used to find the best average recognition score for the whole gestures at different values of k *(best recognition score was at k=1)*. (*)

The following figure reveals the recognition score of each of the seven gestures composing the database. It is clear that there are 4 gestures achieved 100 % recognition score, and 3 gestures achieved 75 % recognition score, making a total average recognition score through all gestures ~ 90%.

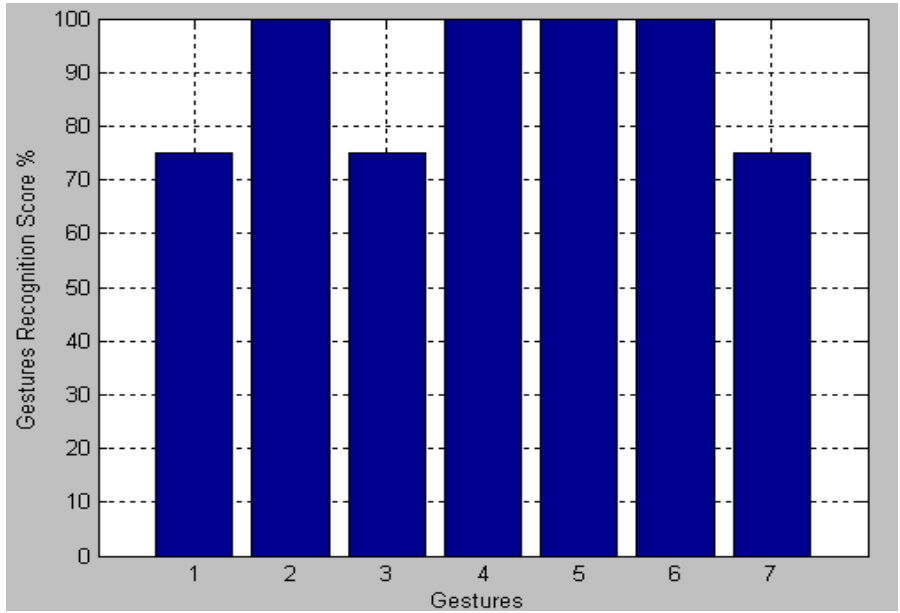

Fig. (1-16) Static gestures recognition score

The previous obtained results using the appearance based characterisation method prove the pertinence of this method in terms of the obtained total average recognition score. However, a better recognition score could be obtained when using the 3D hand model based characterisation method, but this method is more complicated and takes more time for calculations. In general, the previous obtained total recognition score is accepted for the main purpose of this chapter.

---

(*) For more information about the k-nearest neighbor classification method *(KNN)* and the cross validation, please refer to the attached appendixes (5, 6) at the end of this report.

# Chapter 2

## Dynamic gestures recognition

Dynamic gestures are gestures that change in time from the initial position to the final position, and in this work dynamic gestures are classified in two categories:

- *Cyclic gestures*: which are gestures that change from the initial position to an intermediate position, then back to the initial position, i.e. the initial and final positions are the same (ex: Bye bye).
- *Linear gestures*: which are gestures that change linearly from the initial to the final positions (ex. when the arm gets raised and make the hand pointing to the left or to the right).

In the database, cyclic gestures are considered as gestures of small amplitude, i.e. they change dynamically but the amplitude of the dynamic change is small, while linear gestures are, mostly, considered as gestures of large amplitude.

Based on the above classification, two recognition systems are used; one for gestures of small amplitude and the other one for gestures of large amplitude. On the other hand, for a test gesture captured by the camera of the robot, it will be classified first in real time to one of the amplitude categories (large or small) using the optical flow *(see chapter 3)*, then it will be processed by the recognition system corresponding to its amplitude as being low or large.

## (1) Small amplitude gestures recognition

Small amplitude gestures in my database are composed of five cyclic gestures and one linear gesture of nine persons:

| | | |
|---|---|---|
| 1- Bye bye (two hands) | 3- Bye bye (left hand) | 5- Yes |
| 2- Bye bye (right hand) | 4- Stop | 6- No |

Actually, in this work i supposed that the person who will perform the gesture will make it in a spontaneous way, i.e. the amplitude and the speed of the gesture may get different between different persons of different. However, even with that possible variation in the amplitude or the speed of the above mentioned gestures, they will stay considered as small amplitude gestures. So, by taking into consideration the factors of amplitude and speed variation, my idea in analysing these gestures is to extract key frames from the video basing on some criterion will be explained in details in this chapter, then we calculate a relevant characteristic vector and use the k-nearest neighbour for the classification.

### (1-1) Key frames extraction

Key frame extraction is an important technique in video summarization, browsing, searching, and understanding. The extracted key frames contain complex motion and are salient with respect to their neighbouring frames, and can be used to represent actions and activities in the video [21] [22].

Two techniques are used to extract key frames which are intra- frame method and inter-frame method. In this work i used the inter-frame method to extract the required key frames as explained later.

### (1-1-1) Inter-frame Method

This method of key frames extraction enables us to study image histograms that change fast relatively to the surrounding frames which mean that they contain important information. Assume the histograms of a frame and its neighbourhood frame (*x* frames leading or lagging) are *H (i)* and *H (i ± x)* respectively, and each contains *n* bins. The histogram intersection *HI* between two histograms is defined as [23]:

$$HI = \frac{\sum_{i=1}^{n} \min(H(i), H(i \pm x))}{\sum_{i=1}^{n} H(i)} \qquad (1)$$

The denominator normalizes the histogram intersection and makes the value of the histogram intersection between 0 and 1. The *HI* value is actually proportional to the number of pixels from the current frames that have corresponding pixels of the same motion vectors in the neighbourhood frame. A higher *HI* value indicates higher similarity between two frames.

After calculating histogram intersection throughout all video frames, i calculated the local maximum points through the curve *(see fig (1) the black points)*, then in each group of 5 or 6 black points, i chosed the maximal point *(see fig (1) the red points)*. The extracted key frames correspondents to the red points are shown in figure (2).

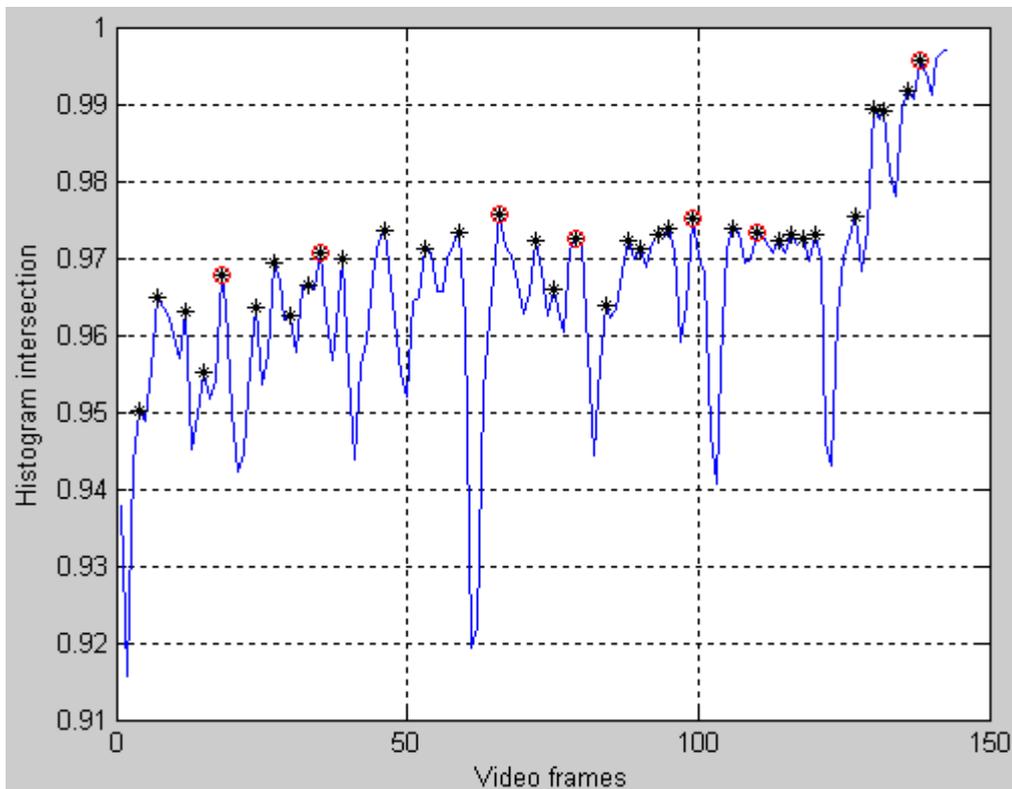

**Fig. (2-1)** Key frames extraction

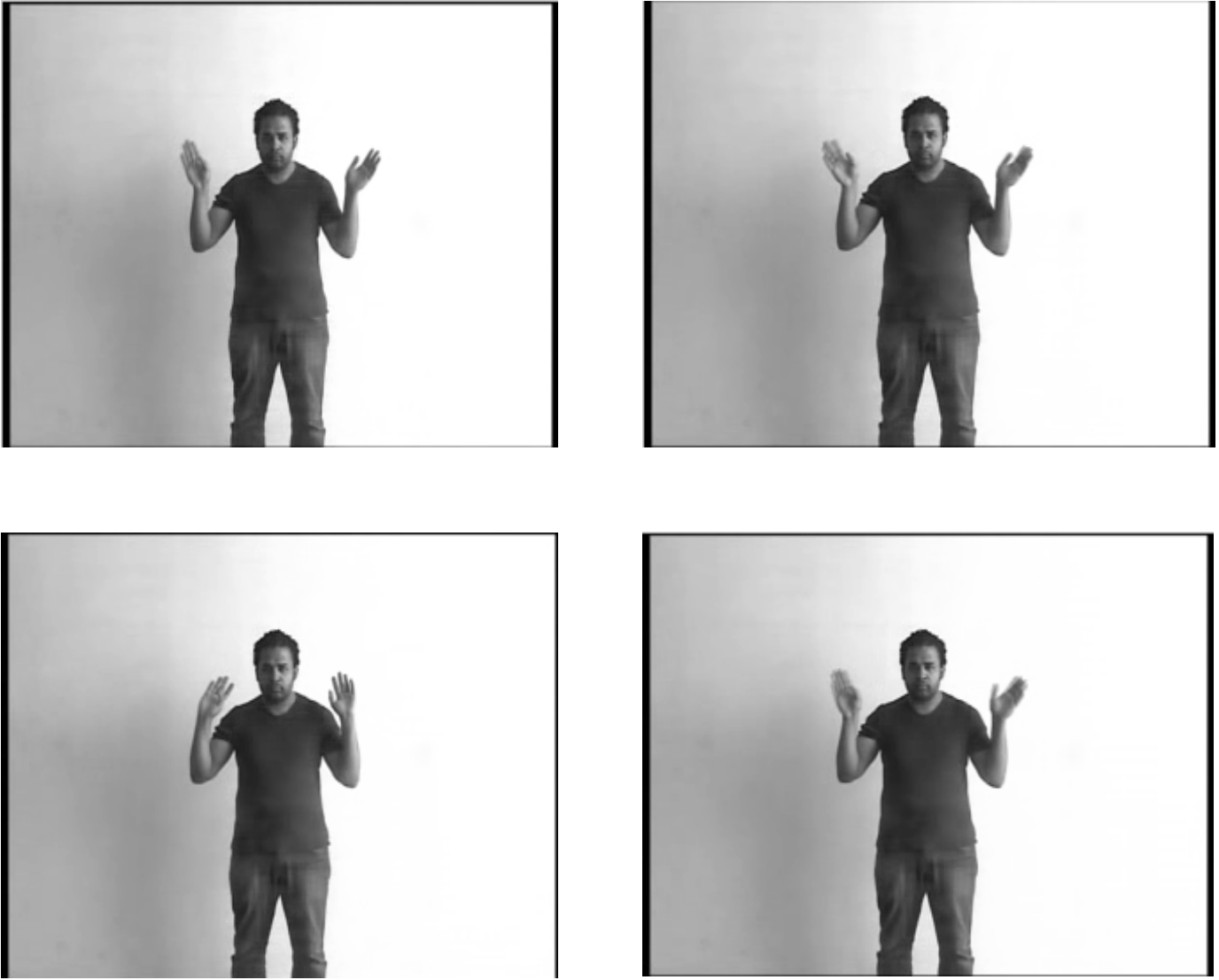

**Fig. (2-2)** Examples of extracted key frames

In the previous extracted key frames, we observe that there are some key frames where the hands are in similar and near positions; i.e. highly correlated and others are weakly correlated. So, from the whole extracted key frames, i choose just two key frames that give the lower possible correlation, and then i start preparing these two key frames in order to calculate the characteristic vector as will be indicated later.

**(1-2) Extracted key frames subtraction**

In order to characterise the gestures made, we have to focus only on the dynamic part of the body, but there will be many factors to take into consideration:

1- Light          2- Clothes color          3- Background

All these factors may affect badly the process of extracting the part of interest of the body while making gestures *(see chapter 1)*. The factor which may affect the process of dynamic gestures recognition in a different way to what was explained in the previous chapter is the "Background" which has to be static. So, it will be effective to subtract the extracted two key frames in order two eliminate all static details, like the background and the static parts of the body, regardless of the skin color or clothes color *( see figure 3)* [24].

The following figure illustrates the idea of key frames subtraction, it is clear that all static parts have been already eliminated , except some few details appear in the image apart from the moving hands like body boundaries. These undesired details could be eliminated by suitable noise filtering, and then this image gets binarized in order to calculate the characteristic vector.

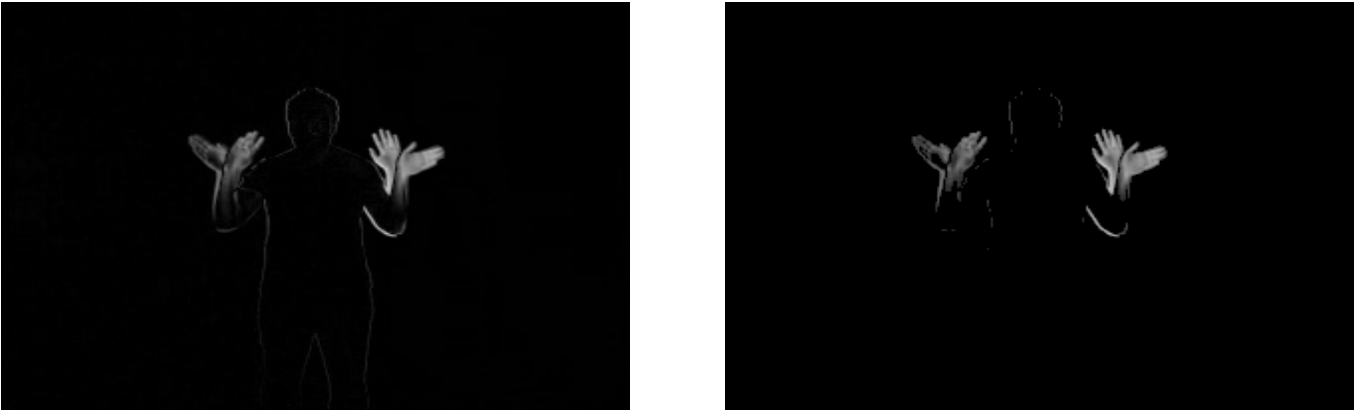

**Fig. (2-3)** Difference image of the two extracted key frames *(before and after filtration)*

### (1-3) Image processing

After obtaining the difference image, it has to be processed to filtrate it from any unnecessary details that may affect badly the calculation of the characteristic vector, these processes are:

- Image thresholding: where the original *RGB* image is converted to black and white image by applying a specific thresholding value over the image, where pixels of values inferior to the threshold value would be considered as black pixels, otherwise, they would be considered as white pixels *(figure 4)*.

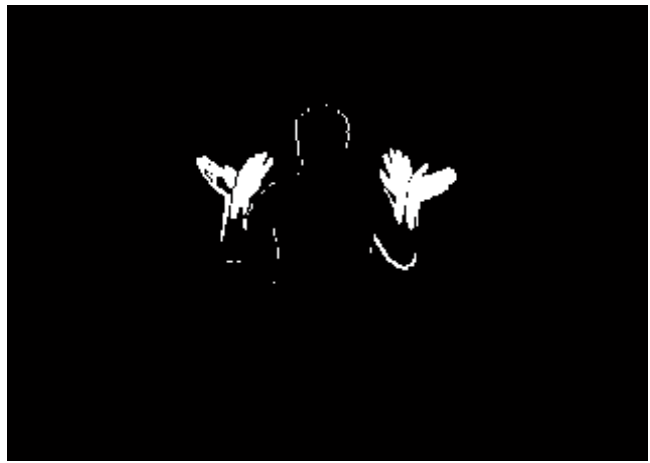

**Fig. (2-4)** Thresholded difference image

- Counting the connected components: Mostly, all the undesired details remaining in the thresholded difference image are of small sizes, i.e. after counting all the connected components and calculating the size of each population, it is possible to precise a threshold size value in order to eliminate all undesired details of small sizes *(Referring to image 4, the bigger populations are the left two connected hands and the right two connected hands which are also the components of concern in this image . It is clear that the sizes of the undesired details are so small with respect to the left and right hands, which facilitate eliminating these small details by a suitable threshold size value)*.

### (1-4) Calculation of the characteristic vector

The characteristic vector has to characterize in a pertinent way the difference image, therefore, this vector is composed of [24] [25]:

- The centre of gravity :
$$\bar{x} = \frac{\sum_{x,y} x|D(x,y)|}{\sum_{x,y}|D(x,y)|} \quad , \quad \bar{y} = \frac{\sum_{x,y} y|D(x,y)|}{\sum_{x,y}|D(x,y)|} \qquad (2)$$

- The mean absolute deviation of a pixel *(x, y)* from the centre of motion $\sigma(t)^T = [\sigma_x(t), \sigma_y(t)]$ where:

$$\sigma_x(t) = \frac{\sum_{x,y}|D(x,y)(x-\bar{x})|}{\sum_{x,y}|D(x,y)|} \quad , \quad \sigma_y(t) = \frac{\sum_{x,y}|D(x,y)(y-\bar{y})|}{\sum_{x,y}|D(x,y)|} \qquad (3)$$

- Intensity of motion :
$$i = \frac{\sum_{x,y}|D(x,y)|}{\sum_{x,y} 1} \qquad (4)$$

- *(In case of gesture Bye bye 2 hands)*, we can add the difference between the centroids of left hands' span and right hands' span *( see figure 5)* to the characteristics list.

So the total characteristic vector is composed of 12 elements and could be presented as follows *in case of gesture Bye bye 2 hands)*:

$$Vector = [\bar{x}_1 \ \bar{y}_1 \ \bar{x}_2 \ \bar{y}_2 \ \sigma x_1 \ \sigma y_1 \ \sigma x_2 \ \sigma y_2 \ abs(\bar{x}_1 - \bar{x}_2) \ abs(\bar{y}_1 - \bar{y}_2) \ i_1 \ i_2] \qquad (5)$$

and the total characteristic vector in all the rest gestures in the database could be presented as follows:

$$Vector = [\bar{x}_1 \ \bar{y}_1 \ 0 \ 0 \ \sigma x_1 \ \sigma y_1 \ 0 \ 0 \ abs(\bar{x}_1 - 0) \ abs(\bar{y}_1 - 0) \ i_1 \ 0] \qquad (6)$$

The following figure indicates the characteristic vector drawn as a white ellipse in the left and right hands, and this is the vector used to construct the database and for classification.

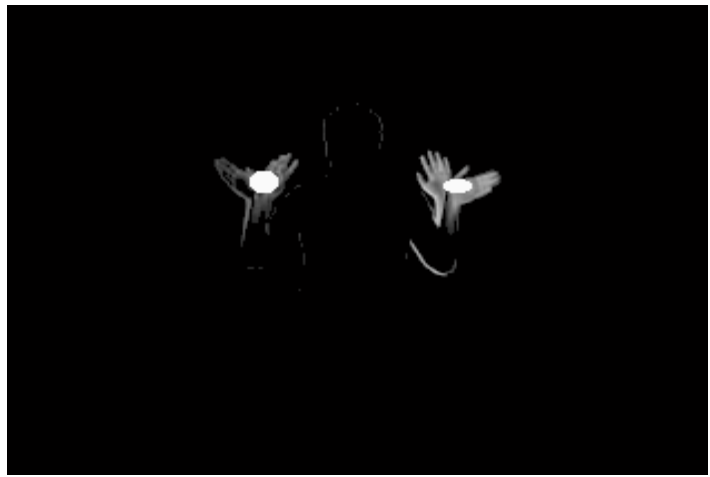

**Fig. (2-5)** Characterizing ellipses of the gesture

## (1-5) Classification

After constructing the database, i used the k- nearest neighbour classification method *(KNN)* supported by the cross validation to find out the best average recognition score for the whole gestures at different values of k. The following figure indicates that the best recognition score was at k=1 (~about 95 %), then decreased with greater values of k, till being constant at k=5 with a recognition score about 83 %.

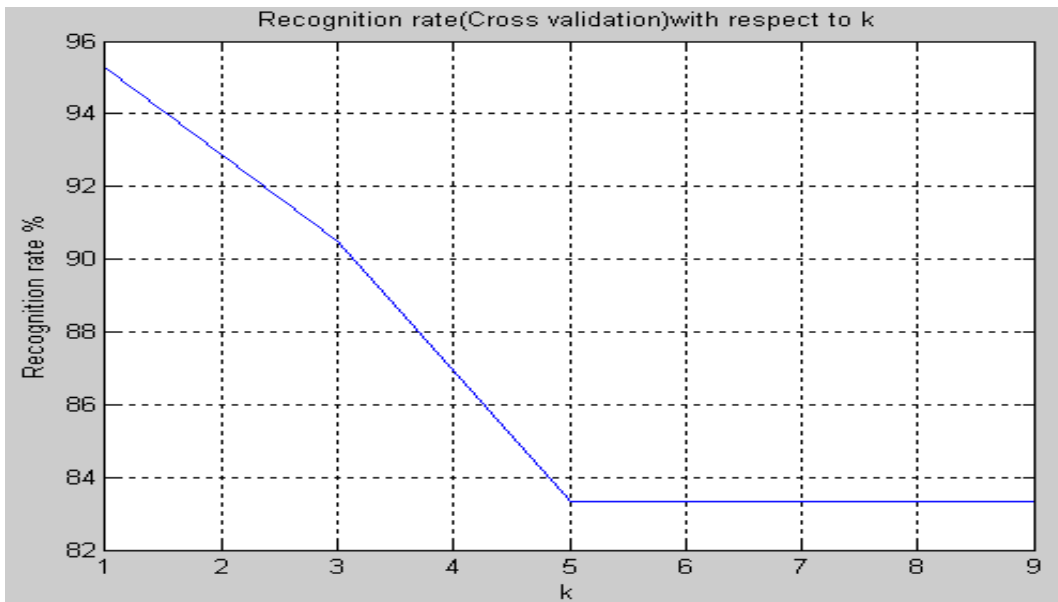

**Fig. (2-6)** Recognition score at different values of k *(S.A gestures)*

The following figure indicates the recognition score of each gesture of the 6 gestures composing the database, it is clear that the recognition score of the first 5 gestures *(mentioned in order in the beginning of this chapter)* are 100% , but the recognition score of the last gesture *(No)* approaches 75 %.

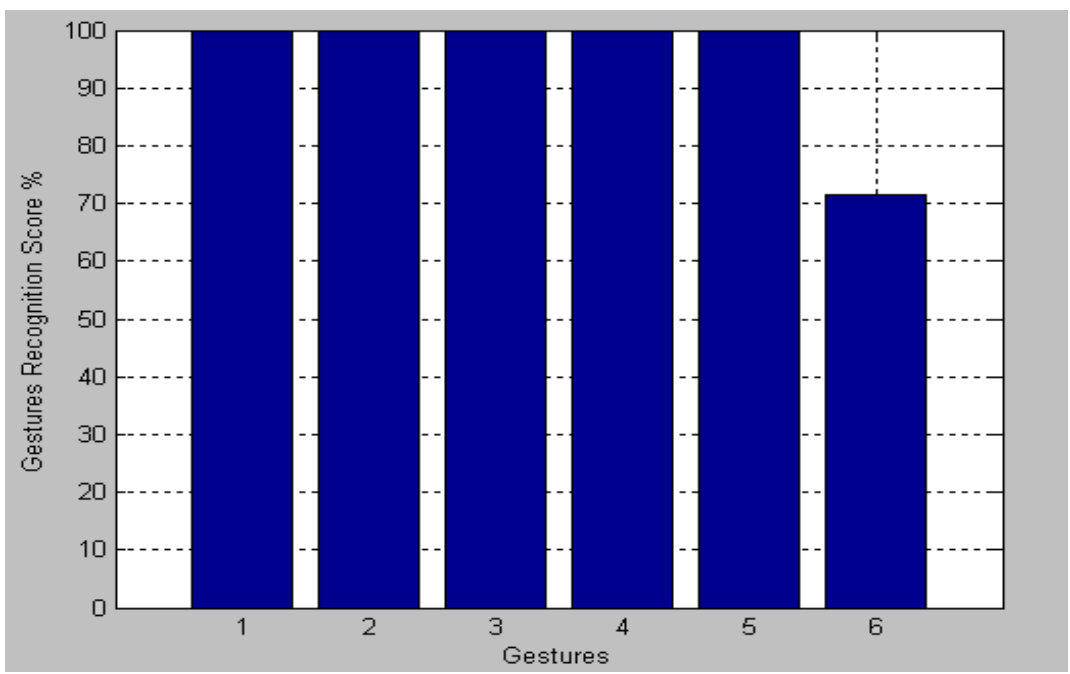

**Fig. (2-7)** Gestures recognition score *(S.A gestures)*

The following figure reveals the confusing gestures for that system of recognition that were misclassified. In this figure, the red line presents the ideal probability of appearance of each gesture, i.e. the frequency of appearance, in the final result classification vector. Gestures 2 and 5 *(Bye bye (right hand), Yes)* appear more frequent in the final result classification vector than other gestures, and as a result, the recognition score of gesture 6 *(NO)* was reduced. In other words, the recognition system confused between gestures 2, 5 and 6, but at the end, the total average recognition score of all gestures still reasonable.

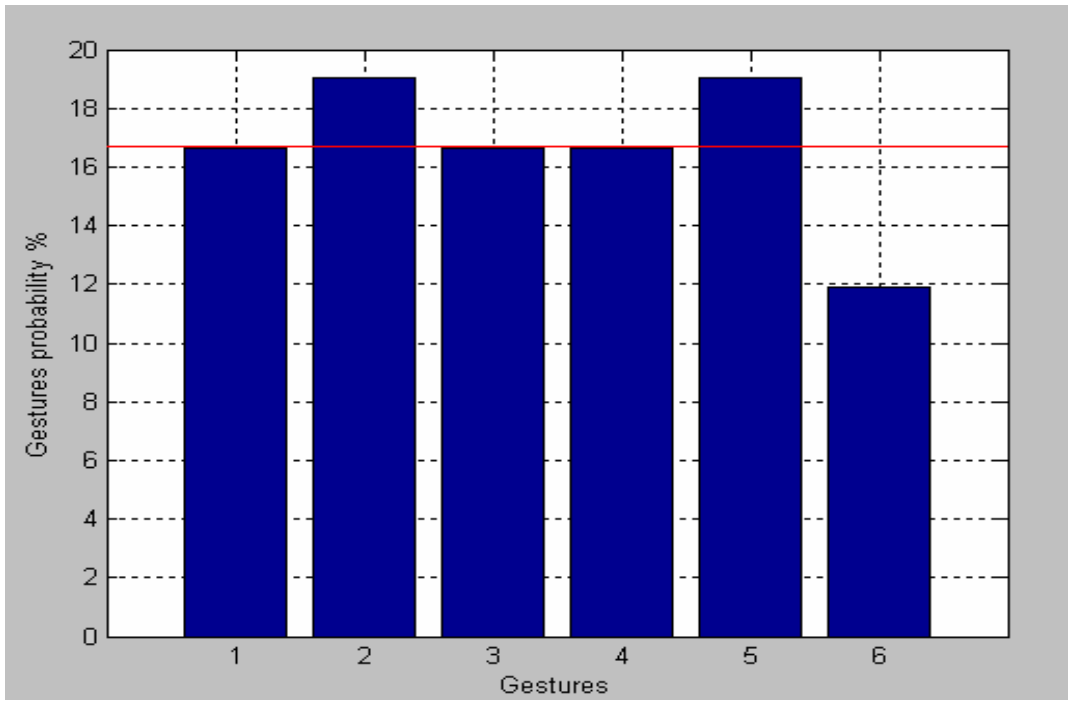

**Fig. (2-8)** Misclassified gestures *(S.A gestures)*

In fact, during the classification and recognition process, there is a special difficulty with gestures 5 and 6 *(Yes, No)*, which is the direction of the gesture with respect to the axis of the robot camera, so if it is bi-axial like gestures from 1 to 4, there will not be any problem in obtaining a pertinent difference image and consequently the classification score will be high *(see figure 7)*.

On the other hand, gestures 5 and 6 are co-axial with the camera axis, so when binarising the image and calculating all the connected objects in order to eliminate the unnecessary details, i found that the size of the head population is small as if it belongs to the unnecessary details. In such cases, in order to overcome this point, i counted first all the connected objects and calculated their sizes, and then i compared these sizes to a size threshold value to decide the probability of the gesture to be co-axial or be-axial gesture *(in bi-axial gestures, the calculated sizes will be more bigger than the threshold value, and vise versa with the co-axial gestures)*.

Once the gesture is classified as being co-axial or bi-axial, i precised another size threshold value to differentiate between gestures 5 and 6 *(Yes and No)*. This new threshold value is obtained by testing many gestures of this type *(co-axial "Yes and No" gestures)* made by many persons in order to calculate this approximate value. Definitely, this problem could be resolved in a more pertinent and effective way if it is possible to study the 3D view of the captured gesture, but the robot vision system is not able to construct 3-D views. So my solution was oriented to resolve this issue as a 2-D view, and i think that the total average recognition score is accepted *(see figure 7)*.

## (2) Large amplitude gestures recognition

Large amplitude gestures in my database are composed of nine linear gestures of nine persons some of them are similar but in opposite directions, therefore, one of the important issues in the classification system, is to be able to well classify those gestures of opposite directions, these gestures are [(*)]:

1- Kowtow
2- Walk from left to right
3- Walk from right to left
4- Right arm up
5- Right arm down
6- Right arm rotation (clock wise)
7- Right arm rotation (counter clock wise)
8- Right arm pointing to right
9- Left arm pointing to left

Large amplitude gestures' recognition system includes the same processes as in the small amplitude gestures' recognition system like: obtaining the difference image, eliminating unnecessary details ….etc. But, in place of making calculations only for one state as in the small amplitude gestures' recognition system, they will be done for 5 consequent states characterising each of the large amplitude gestures.

The reason behind analysing large amplitude gestures in consequent frames *(states)* is that as long as these gestures will be performed spontaneously; there must be attention for the differences in their amplitude and speed according to the person who performs them. So, suppose that we try to process large amplitude gestures by extracting just two key frames as in the small amplitude gestures, we will find that the characterising ellipse of one person is more smaller or bigger than the characterising ellipse of another one doing the same gesture *(because of the difference in speed as mentioned before which affects the extracted key frames)*, therefore the recognition result may be badly affected.

In order to avoid that, it will be more effective to analyse large amplitude gestures via consequent states to minimize the possible bad effects of speed and amplitude differences, therefore, the characterising vectors of one gesture made by many persons *(each vector is a multi states vector)* will be more comparable to each other, consequently, recognition score will be more pertinent.

This characterization vector *(according to the classification results discussed later)* is reasonably pertinent in describing any gesture without confusing it with other gestures even in opposite directions because its elements include mainly the centroid and other elements characterising the white ellipse. So, if we want to classify gestures 2 and 3 *(motion from left to right and vise versa)*, there won't be any problem because the transition of states' centroides in the two gestures will be distinguished and in opposite directions. Therefore, the recognition score will be in the accepted range *(see figure 11)*.

The following figure indicates the idea of consequent states characterisation in order to well recognize and classify large amplitude gestures. The right column frames are the consequent states of me moving from right to left, while the left column frames are the consequent states after being characterised each. Therefore the final characterisation vector for this movement will be composed of 4 included characterisation vectors *(the white ellipses)* presenting the consequent states within the movement [24] [25].

---

[(*)] Gestures 2, 3 and 5 are included in the database just to clarify the idea of dynamic gestures' characterization and classification, however; they will not be used in any interactional scenarios because they don't convey any communicating messages. For more details, please refer to appendix (2) at the end of this report.

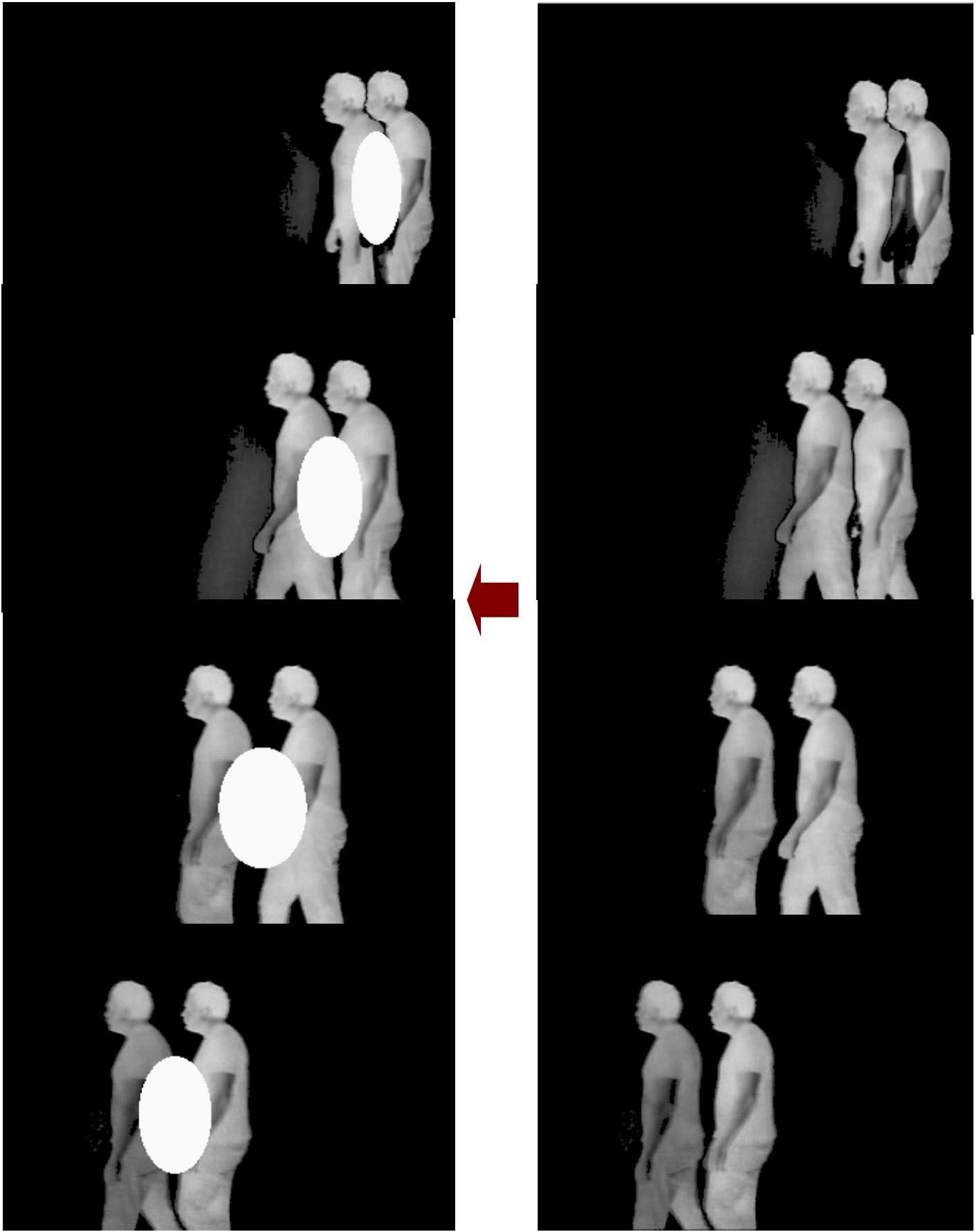

**Fig. (2-9)** Motion consequent states characterization

## (2-1) Classification

In order to classify and recognize consequent states gestures, there are 2 main techniques to handle these sequences:

- *Sequence based classification*: where the distance function which measures the similarity between sequences determines the quality of the classification significantly, i.e. the *KNN* method applied in the classification system of small amplitude gestures.
- *Model based classification*: such as using *Hidden Markov Models (Hmm)*.

It is known about *Hmm* classification method to be more precise and effective than *KNN* method in sequence classification. But, the pertinence and relevance of the characteristic vector made the results of the k-nearest neighbour method (*KNN*) highly reliable. So, i decided to generalize it as the classification method used in the system *(supported by the cross validation)*. Besides, it is more easy and direct than the *Hmm* method.

The following figure indicates that the best recognition score was at k=1 (~about 92.6 %), then decreased with greater values of k, till reaching about 85 % recognition score at k=9.

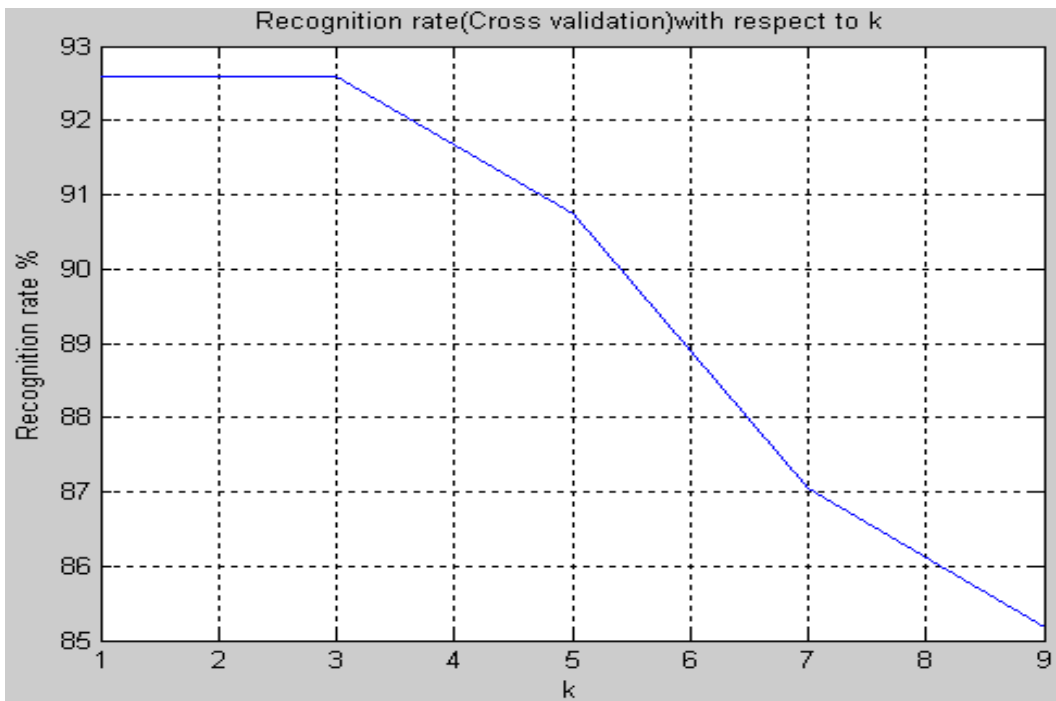

**Fig. (2-10)** Recognition score at different values of k *(L.A gestures)*

Figure (11) indicates the recognition score of each gesture of the 9 gestures composing the database. It is clear that there are 5 gestures at score 100 % while the rest 4 gestures are at score 83 %. The partially misclassified gestures are: *Walk from left to right, Right arm rotation (clock wise), Right arm rotation (counter clock wise) and Right arm pointing to right*. However, the total average recognition score ~92.6 % is a good score.

Figure (12) reveals the confusing gestures and the misclassified gestures as a result. We observe that the appearance of gesture 4 *(Right arm up)* in the final result classification vector was excessive, i.e. gesture 4 affects negatively the recognition score of other gestures.

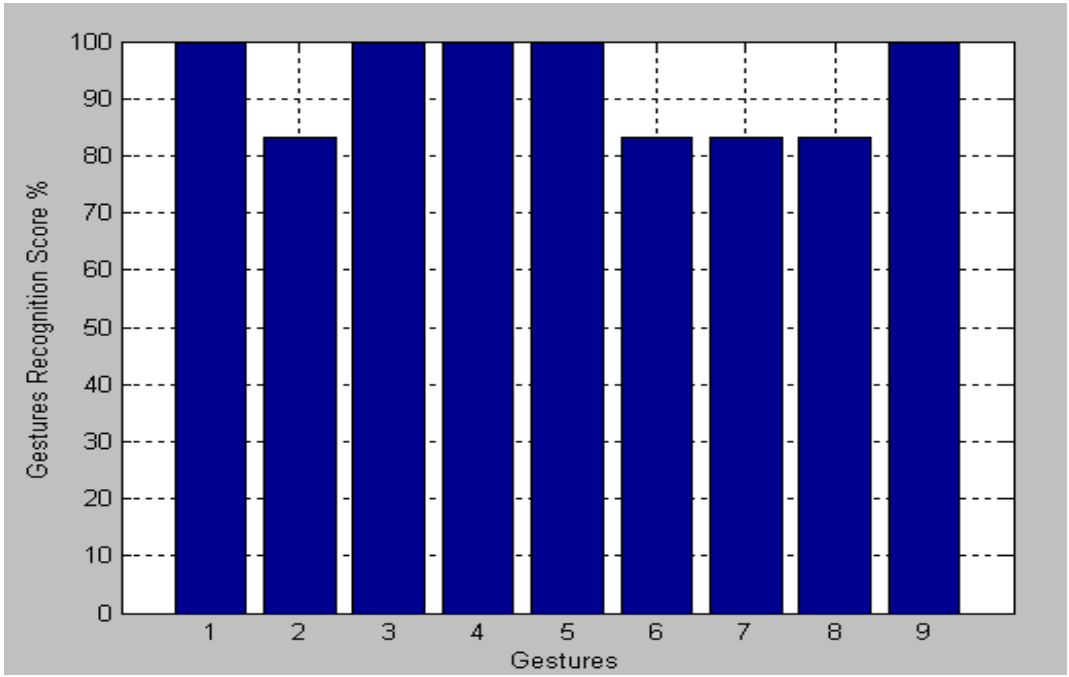

**Fig. (2-11)** Gestures recognition score *(L.A gestures)*

Besides, in the following figure, we can see also that, however the appearance of gestures 6 and 8 labels in the final classification vector is in the ideal range *(the red line)*, the recognition score of these gestures is not at 100 %, i.e. they appeared frequently in the final result classification vector, but just as misclassified labels *(i.e. in wrong positions in the result classification vector)*.

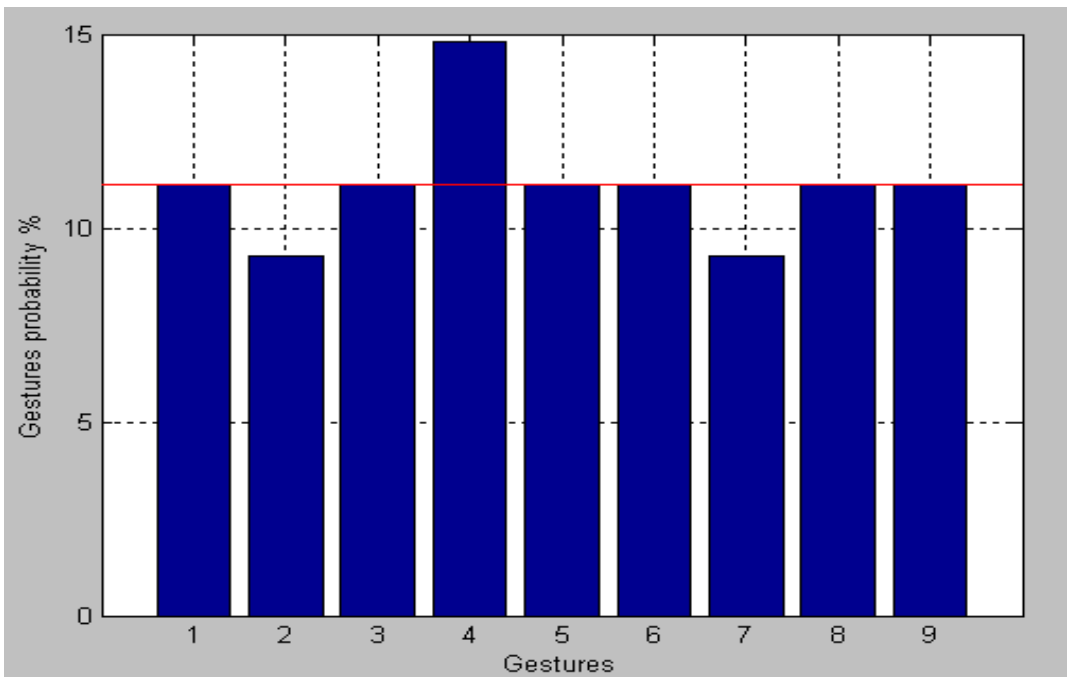

**Fig. (2-12)** Misclassified gestures *(L.A gestures)*

Last but not least, the classification process could be summarized as in the following figure, where the feature database is constructed by the characteristic vectors of the training gestures. This database will be compared to a test gesture's characteristic vector in order to classify it.

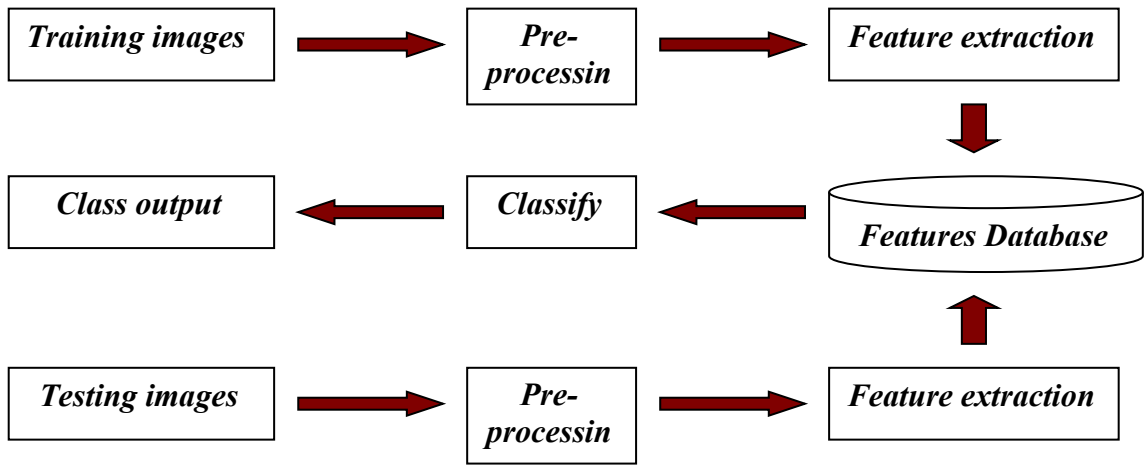

**Fig. (2-13)** Steps of the classification process

The classification and recognition of small and large amplitude gestures had shown reasonable results with average total recognition score in both ~94 %. So, the following step is to find a possibility to differentiate between small and large amplitude gestures in real time during acquisition by the robot camera to orient the coming test gesture towards one of the two used classification programs (*small amplitude gestures' or large amplitude gestures' classification programs which are of the same type KNN but different in the way they handle the database*). The technique used to differentiate between small and large amplitude gestures in real time is called "Optical flow" and is discussed clearly in the coming chapter.

# Chapter 3

## Optical flow

In the last chapter, i focused on the classification system used to recognize small and large amplitude gestures. The classification is based, in both cases, on the k-nearest neighbour algorithm (*KNN*), but the way it deals with the database in both cases is different in a way that assures a good total average recognition score. Therefore, in order to get the optimum classification performance, it is necessary to differentiate between small and large amplitude gestures in real time during acquisition and to know which classification program to use with the test video *(as indicated in figure 1)*. For that, the "*optical flow*" is used.

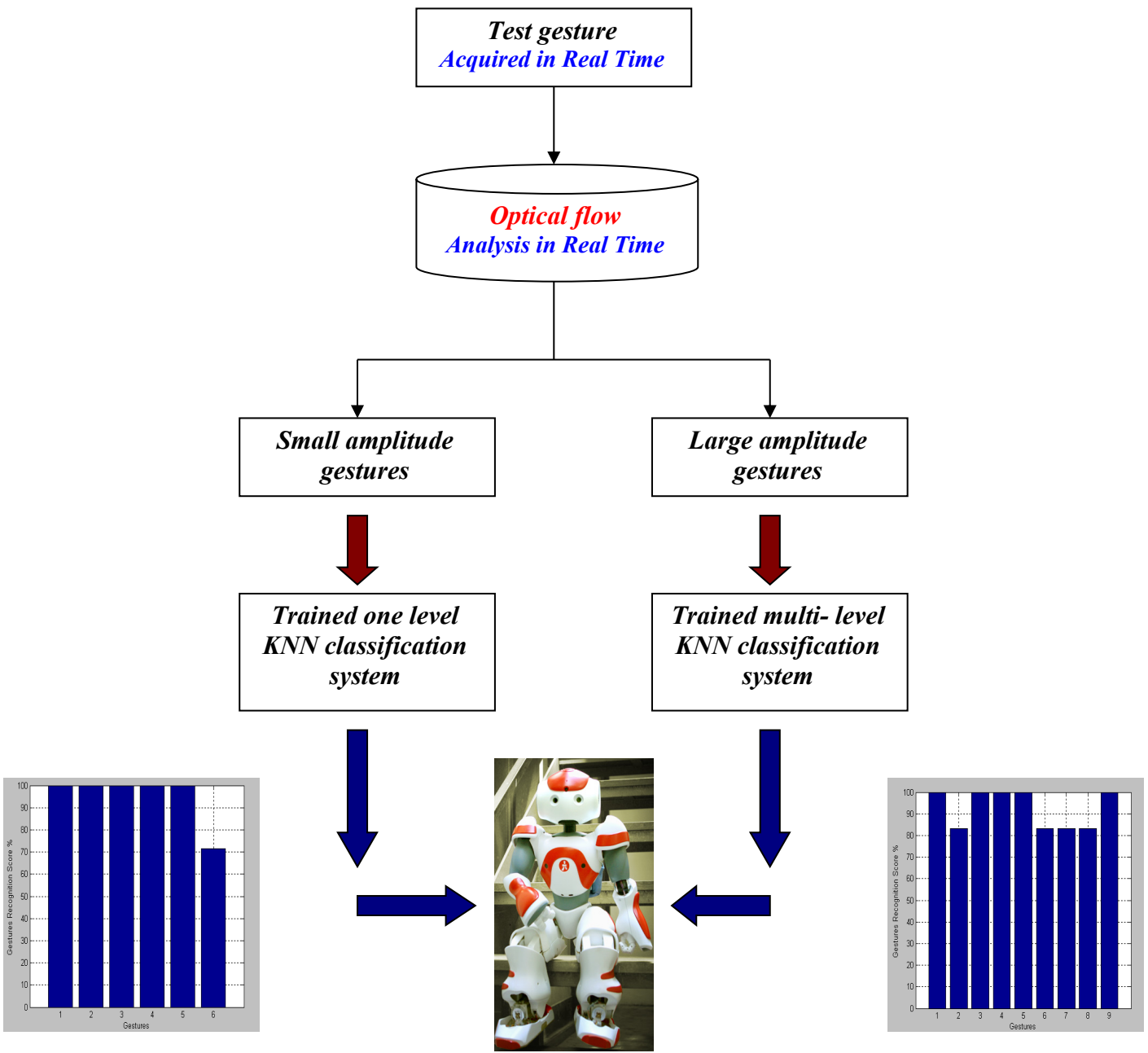

**Fig. (3-1)** Optical flow and robot control

The optical flow [31] [32] [33] is the pattern of apparent motion of objects, surfaces, and edges in a visual scene caused by the relative motion between an observer (an eye or a camera) and the scene, in other words, it is the velocity field that wraps one image to another. Based on that, i constructed a *MATLAB Simulink* model, that acquires the gesture via the camera of the robot in real time and then tracks the moving parts in the scene using the optical flow. This tracking will be useful in detecting the centroid of these moving parts in each frame acquired by the camera. When comparing the detected centroides' values throughout the acquired frames, it will be easy to have an idea about the amplitude of the gesture as being small or large by calculating the transition of the centroid from the first frame to the last frame. The following figure depicts the optical flow field of a ball rotating from left to right.

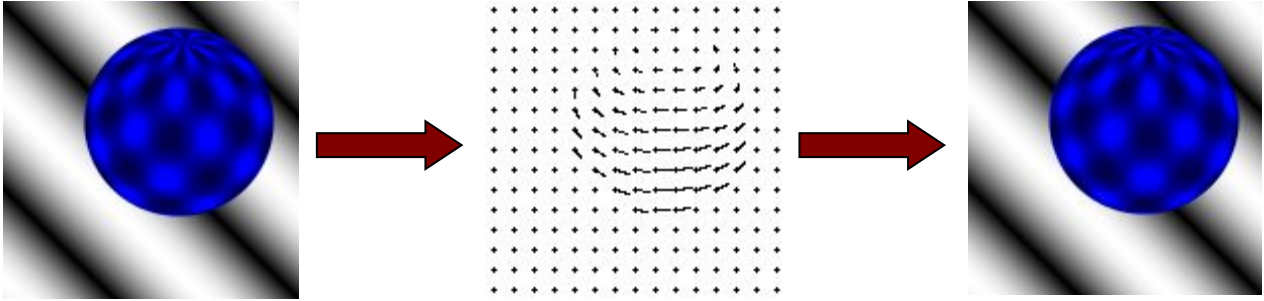

**Fig. (3-2)** Optical flow field

# (1) Estimation of the optical flow

In order to estimate the optical flow between two images, the following optical flow constraint equation should be solved:

$$I_x u + I_y v + I_t = 0 \tag{1}$$

Where:
- $I_x, I_y$ and $I_t$ are the spatiotemporal image derivatives.
- $u$ is the horizontal optical flow *(velocity)*.
- $v$ is the vertical optical flow *(velocity)*.

Two methods are mainly used to solve this equation to estimate $u$ and $v$: *Horn-Schunck* method and *Lucas-Kanade* method. In this work, i concentrated mainly on the first method; *Horn-Schunck method*.

## (1-2) Horn-Schunck method

By assuming that the optical flow is smooth over the entire image, the *Horn-Schunck* method computes an estimate of the velocity field $[u\ v]^T$, that minimizes the following equation [32]:

$$E = \iint (I_x u + I_y v + I_t)^2 dxdy + \alpha \iint \left\{ \left(\frac{du}{dx}\right)^2 + \left(\frac{du}{dy}\right)^2 + \left(\frac{dv}{dx}\right)^2 + \left(\frac{dv}{dy}\right)^2 \right\} dxdy \tag{2}$$

Where:
- *E* is the global energy of the optical flow to be optimised by minimizing all possible distortions in the flow, for this reason this method is highly dependant on the smoothness of the flow.

- $\dfrac{du}{dx}$ and $\dfrac{du}{dy}$ are the spatial derivatives of the optical velocity component $u$.
- $\alpha$ Smoothness factor *(positive scalar constant)*: if the relative motion between two images or video frames is large, the value of the smoothness factor should be sufficiently large and vise versa.

The *Horn-Schunck* method minimizes the previous equation to obtain the velocity field, [u v], for each pixel in the image, which is presented by the following equations:

$$u_{x,y}^{k+1} = \overline{u}_{x,y}^{-k} - \frac{I_x\left[I_x \overline{u}_{x,y}^{-k} + I_y \overline{v}_{x,y}^{-k} + I_t\right]}{\alpha^2 + I_x^2 + I_y^2} \qquad (3)$$

$$v_{x,y}^{k+1} = \overline{v}_{x,y}^{-k} - \frac{I_y\left[I_x \overline{u}_{x,y}^{-k} + I_y \overline{v}_{x,y}^{-k} + I_t\right]}{\alpha^2 + I_x^2 + I_y^2} \qquad (4)$$

In these equations, [$u_{x,y}^k$ $v_{x,y}^k$] is the velocity estimate for a pixel "A" at (x, y) at iteration k, while [$\overline{u}_{x,y}^{-k}$ $\overline{v}_{x,y}^{-k}$] is the neighbourhood average of [$u_{x,y}^k$ $v_{x,y}^k$]. So, it is clear that this method is an iterative method to calculate the optical flow between consecutive video frames *(the superscript k+1 denotes the next iteration to be calculated and k is the last calculated result)*.

## (2) Tracking of the dynamic parts of the body using the optical flow

Based on the principal of the optical flow explained above, my *Simulink* model acquires in real time the video *(frame by frame)* captured by the robot camera via *TCP* socket communication then starts a real time image processing.

It calculates the optical flow between consequent frames in terms of velocity, and then applies a specific velocity threshold on the calculated optical flow to exclude the static parts of the body. The extracted dynamic part of the body, in binary form, will go through a median filter and a morphological closing operation to ensure the optimal elimination of the undesired details, then the connected regions in the binary image will be counted, and each of these regions will be surrounded by a green rectangle according to the size *(so it is so clear the importance of these image processing operations and how they have to be precise , because if some noise still in the final binary image , it  may be surrounded by a green rectangle as a moving object and that may make error in the calculation of the gesture's amplitude. So, it is necessary to study well the environment in which the gesture is made and to observe the effect of all possible noise sources like lights on the tracking in order to make what is needed to eliminate that effect if necessary)*.

*Simulink image toolbox* gives the possibility to analyse the surrounded region by the rectangle statistically in terms of its area, centroid, orientation and .etc. When comparing the obtained centroid data throughout all the acquired frames, it will be easy to fix a centroid transition threshold to differentiate between small and large amplitude gestures.

The following figure reveals the moving parts of my body in 4 different dynamic gestures: *Bye bye two hands, Bye bye right hand, left arm pointing, and right arm down*, where all the detected dynamic parts are surrounded by green rectangles that will be analysed to get the centroid values necessary for the gesture's amplitude calculation.

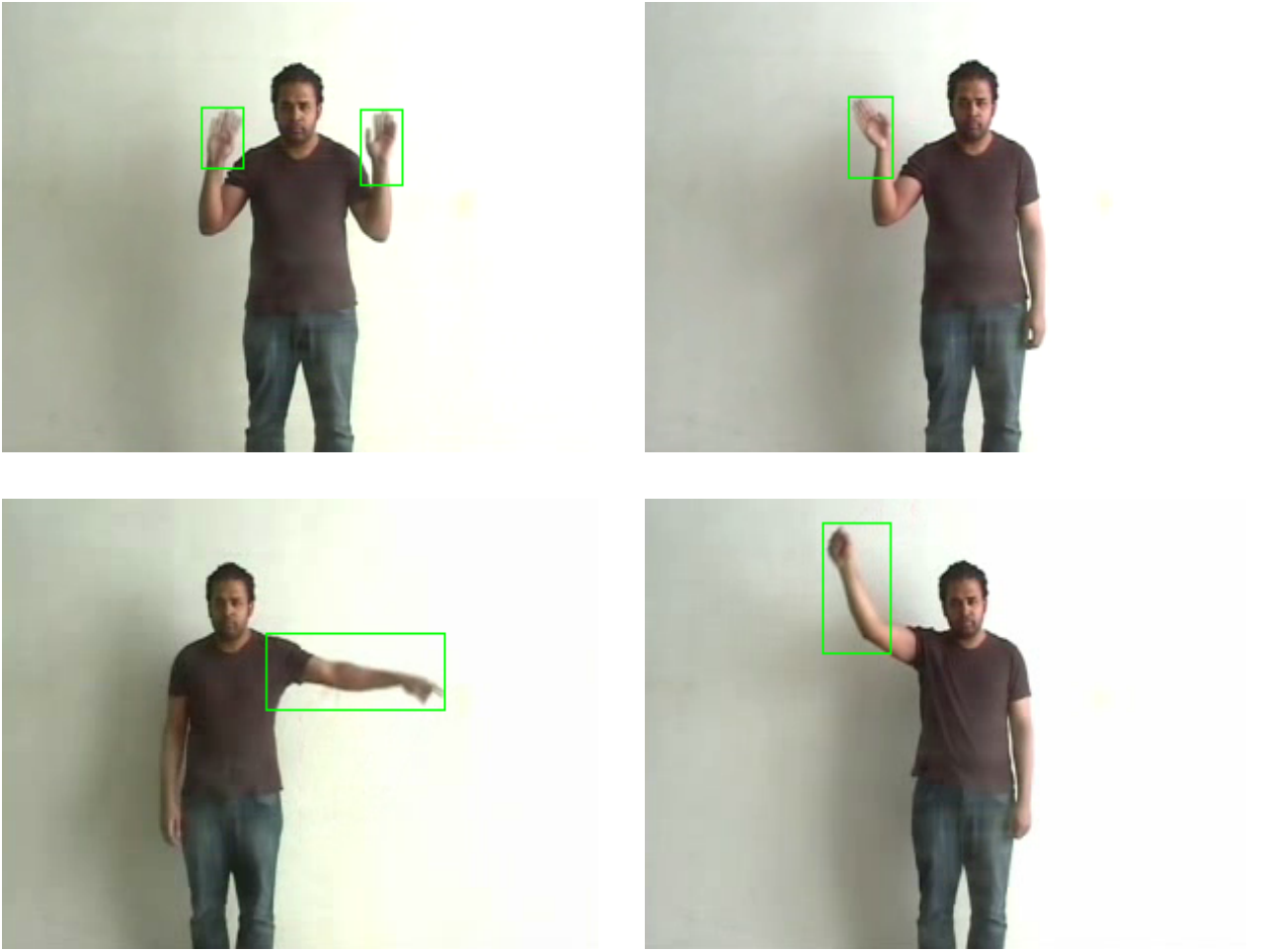

**Fig. (3-3)** Dynamic body parts tracking

## (3) Examples on the gesture's amplitude calculation

As referred before, the amplitude of a gesture will be calculated by the transition of the rectangle's centroid from the first frame to the final frame. So, the total transition Euclidian distance could be summarized by the following equation:

$$\text{Centroid\_Transition} = \sqrt{(x_{finalframe} - x_{firstframe})^2 + (y_{finalframe} - y_{firstframe})^2} \quad (5)$$

So for example, in the gesture *Bye bye right hand (small amplitude gesture)*, the centroid transition value is:

$$\text{Centroid\_Transition} = \sqrt{(133.0188 - 127.9951)^2 + (83.4235 - 91.7437)^2} = \mathbf{9.7192}$$

Similarly, in the gesture *left arm pointing (Large amplitude gesture)*, the centroid transition value is:

$$\text{Centroid\_Transition} = \sqrt{(202.4838 - 156.5291)^2 + (93.6265 - 167.0471)^2} = \mathbf{85.6165}$$

Last but not least, in the gesture *right arm down (Large amplitude gesture)*, the centroid transition is:

$$\text{Centroid\_Transition} = \sqrt{(128.6642-131.5066)^2 + (127.8271-39.8289)^2} = \mathbf{88.0441}$$

According to the previous results, the difference in the amplitude between the small and large amplitude gestures is confirmed to be so clear. Consequently, it is so easy to precise a general amplitude threshold level to differentiate between the two classes of gestures, and that was the main target of this chapter.

# **<u>Conclusion</u>**

This research casts light on imitating gestures with a mobile robot in order to be used in ameliorating the social interactional behavior of autistic children, who had shown better interaction with robots than with humans. The main objective is to make the robot able to understand the meaning and the kinematics of any gesture in order to imitate it to encourage the autistic child to interact in an exciting scenario *(bilateral "robot - child" or trilateral "robot- doctor-child" imitational game- see appendix 4)*.

The importance of using social robots to interact with autistic children comes from the statistics which reveal that there is about 3 to 4 autistic children in every 1000 child worldwide, i.e. in a range of one million child, the expected number of autistic children will be increased to reach 3000 to 4000 child. These statistics reveal how important is autism, and how noble and necessary to support autistic children to live and interact in a better way.

As indicated in appendix (3), it is not known for the moment the real causes of autism, which definitely increases the difficulty of treating it medically in a complete way. Therefore, this research tried to explore the potential of embodied communication and user modeling as a tool for training and facilitating social and communicative behavior in children with ASD.

This report focused on two main researching axes: static gestures recognition and dynamic gestures recognition. The first axe *"static gestures recognition"* had illustrated the main problems facing the identification of a hand making gesture and extracting it from an image. These problems could be summarized as:

  1- Background color     2- Light effect     3- Clothes color

The most important factor between the above mentioned factors is the background color which has to be highly contrasted to the color of skin to facilitate separating between background and the foreground pixels. Afterwards, the hand gets extracted and characterized it terms of its contour geometrical parameters like the area, perimeter, and the moments. The k-nearest neighbor classification method *(KNN)* is then used to classify and recognize testing gestures supported by the cross validation to get a total average recognition score of about 90 % which could be considered as an accepted recognition score.

The second researching axe *"dynamic gestures recognition"* had focused on the classification of dynamic gestures basing on their dynamic characteristics like the centroid and the amplitude of motion etc. These gestures could be classified into two categories: low and large amplitude gestures.

Low amplitude gestures, are the gestures that have small dynamic amplitude like: nodding the head (yes or no) or waving with the hand. In order to process these gestures, suitable and relevant key frames should be extracted by which the span between the initial and final position of the hand *(or head)* will be clear in the corresponding difference image and characterized by a white ellipse.

On the other hand, large amplitude gestures are processed in a different way based on dividing the path of gesture into several parts and extracting the corresponding key frames *(eg 6 parts with 6 equivalent key frames)*, then consecutive key frames will be subtracted from each other and then the motion during the gesture will be characterized by a sequence of white ellipses presenting the motion from the initial to the final position.

Despite that hidden markov model *(HMM)* is thought to be the best classification and recognition method in case of sequences, but also the k-nearest neighbor classification method *(KNN)* had approved its pertinence and gave a total average recognition score for the whole small and large amplitude gestures of about ~94% which is a good score, therefore, i decided to use k-nearest neighbor as a general classification method through all gestures of small and large amplitude.

In fact, the reason behind having this high recognition score is the pertinent and relevant characterization vector presented by the centroid and the diameter of the white ellipses, and these vectors will construct the database used in the classification and validation process.

From the above explanation, it is clear that two classification systems of the same type *(KNN)* are used, however, they are different in the way of handling the database; one for the small amplitude gestures and the other system is for the large amplitude gestures. In order to decide for a coming testing gesture captured by the robot camera, which classification system to use, the *"optical flow based Simulink model"* will be useful as it detects the moving part of the body and surrounds it by a rectangle of a tracked centroid coordinates through frames. So, comparing these coordinates of the rectangles in the first and final frames and calculating the Euclidian distance will be a good indicator about the gesture as being of low or large amplitude. That was a general outline for the work done in this research.

# Appendix 1

## Robot NAO

The coming figure presents all the articulations of the robot, these articulations can support up to 25 degrees of freedom, in a simulation for the human body. This robot is composed of programmable motorized modules to make any posture or gesture in the range of its articulations' freedom.

*NAO* robot has the following characteristics:

- x86 AMD Geode 500 MHz CPU
- Wi-Fi 802.11b and Ethernet port
- Programming languages : C, C++ , Python
- Programming software: Aldebaran Choregraph (included), Gostai Urbi Studio (not included) and Microsoft Robotics Studio (not included).

256 MB SDRAM / 2 GB Flash memory
2x 30 FPS CMOS video cam res. 640x480

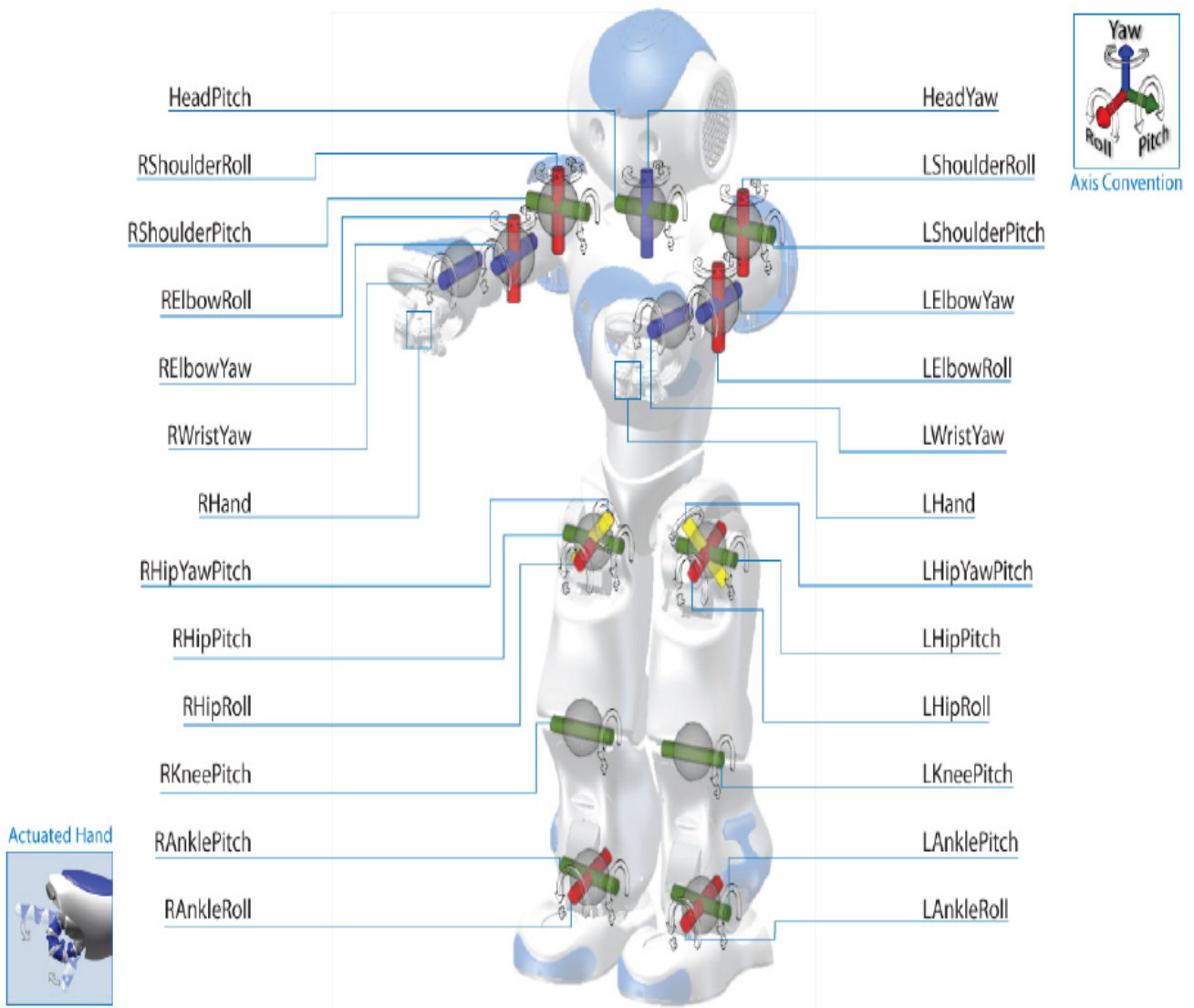

**Fig. (A1-1)** *NAO* robot articulations

The coming table indicates the motion range of all body parts; we can see how wide is the mechanical freedom range which makes the motion of the robot very similar to the same motion made by a human.

| PART | JOINT NAME | MOTION | RANGE (degrees) |
|---|---|---|---|
| Head | HeadYaw | Head joint twist (Z) | -120 to 120 |
| | HeadPitch | Head joint front & back (Y) | -39 / 30 |
| Left arm | LShoulderPitch | Left shoulder joint front & back (Y) | -120 to 120 |
| | LShoulderRoll | Left shoulder joint right & left (Z) | 0 to 95 |
| | LElbowRoll | Left shoulder joint twist (X) | -90 / 0 |
| | LElbowYaw | Left elbow joint (Z) | -120 / 120 |
| | LWristYaw | Left wrist joint twist (X) | -105 to 105 |
| | LHand | Left hand | open & close |
| Left leg | LHipYawPitch | Left hip joint twist (Z45°) | -44 / 68 |
| | LHipPitch | Left hip joint front and back (Y) | -104.5 / 28.5 |
| | LHipRoll | Left hip joint right & left (X) | -25 to 45 |
| | LKneePitch | Left knee joint (Y) | -5 / 125 |
| | LAnklePitch | Left ankle joint front & back (Y) | -70.5 / 54 |
| | LAnkleRoll | Left ankle joint right & left (X) | -45 to 25 |
| Right leg | RHipYawPitch | Right hip joint twist (Z45°) | -68 / 44 |
| | RHipPitch | Right hip joint front and back (Y) | -104.5 / 28.5 |
| | RHipRoll | Right hip joint right & left (X) | -45 to 25 |
| | RKneePitch | Right knee joint (Y) | -5 / 125 |
| | RAnklePitch | Right ankle joint front & back (Y) | -70.5 / 54 |
| | RAnkleRoll | Right ankle right & left (X) | -25 to 45 |
| Right arm | RShoulderPitch | Right shoulder joint front & back (Y) | -120 to 120 |
| | RShoulderRoll | Right shoulder joint right & left (Z) | -95 to 0 |
| | RElbowRoll | Right shoulder joint twist (X) | 0 / 90 |
| | RElbowYaw | Right elbow joint (Z) | -120 / 120 |
| | RWristYaw | Right wrist joint twist (X) | -105 to 105 |
| | RHand | Right hand | open & close |

Table. (A1-1) *NAO* robot motion range

Table (2) reveals the general characteristics of the robot *NAO*; the physical characteristics, leds, energy, processor, networking …etc. Whereas, table (3) indicates the characteristics of the robot motors including the torque, nominal speed and other characteristics.

| Body caracteristics | |
|---|---|
| Height | ~ 58 cm |
| Weight | ~ 4.3 Kg |
| Body type | Technical plastic |
| **Energy** | |
| Charger | AC 90-230 volts/DC 24 volts |
| Battery capacity | ~ 90 min. autonomy |
| **Degrees of freedom** | |
| Head | 2 DOF |
| Arm | 5 DOF in each arm |
| Pelvis | 1 DOF |
| Leg | 5 DOF in each leg |
| Hand | 1 DOF in each hand |
| **Multimedia** | |
| Speakers | 2 Loudspeakers |
| Microphones | 4 Microphones |
| Vision | 2 CMOS digital cameras |
| **Network access** | |
| Connections type | Wi-Fi (IEE 802.11g) |
| | Ethernet connection |

| Actuators | |
|---|---|
| Aldebaran Robotics™ original design based on: | Hall effect sensors |
| | dsPICS microcontrollers |
| | Coreless MAXON DC motors |
| **Sensors** | |
| Different type | 32 x Hall effect sensors |
| | 1 x gyrometer 2 axis |
| | 1 x accelerometer 3 axis |
| | 2 x bumpers |
| | 2 channel sonar |
| | 2 x I/R |
| | Tactile sensor |
| **LED** | |
| Tactile sensor | 12 LED 16 Blue levels |
| Eyes | 2 x 8 LED RGB Fullcolour |
| Ears | 2 x 10 LED 16 Blue levels |
| Torso | 1 LED RGB Fullcolour |
| Feet | 2 x 1 LED RGB Fullcolour |
| **Motherboard** | |
| x86 AMD GEODE 500MHz CPU | 256 MB SDRAM / 2 GB flash memory |
| **Embedded Software** | |
| OS | Embedded Linux (32 bit x86 ELF) using custom OpenEmbedded based distribution |
| Programming languages | C, C++, Urbi script, Python |

**Table. (A1-2)** *NAO* robot general characteristics

| Motor Type 1 | |
|---|---|
| No Load Speed | 8000 RPM |
| Stall Torque | 59.5 mNm |
| Nominal Speed | 6330 RPM |
| Nominal Torque | 12.3 mNm |
| **Reduction ratio type 1** | **201,3** |
| No Load Speed | 238.45 °/s (4.76°/20ms) |
| Stall Torque | 11.97 Nm (without the ratio efficiency) |
| Nominal Speed | 188.67 °/s (3.77°/20ms) |
| Nominal Torque | 2.47 Nm (without the ratio efficiency) |
| **Reduction ratio type 2** | **130,85** |
| No Load Speed | 366.83 °/s (7.33°/20ms) |
| Stall Torque | 7.78 Nm (without the ratio efficiency) |
| Nominal Speed | 290.25 °/s (5.80°/20ms) |
| Nominal Torque | 1.61 Nm (without the ratio efficiency) |

| Motor Type 2 | |
|---|---|
| No Load Speed | 11900 RPM |
| Stall Torque | 15.1 mNm |
| Nominal Speed | 8810 RPM |
| Nominal Torque | 3.84 mNm |
| **Reduction ratio type 1** | **150,27** |
| No Load Speed | 473.72 °/s (9.47°/20ms) |
| Stall Torque | 2.27 Nm (without the ratio efficiency) |
| Nominal Speed | 351.77 °/s (7.03°/20ms) |
| Nominal Torque | 0.57 Nm (without the ratio efficiency) |
| **Reduction ratio type 2** | **173,22** |
| No Load Speed | 412.19 °/s (8.24°/20ms) |
| Stall Torque | 2.61 Nm (without the ratio efficiency) |
| Nominal Speed | 305.16 °/s (6.10°/20ms) |
| Nominal Torque | 0.66 Nm (without the ratio efficiency) |

**Table. (A1-3)** *NAO* robot motors' general characteristics

# Appendix 2

## Gestures

Gestures are considered as a form of non-verbal communication in which visible body actions communicate particular messages using the movement of hands, face, or other parts of the body. One of the main challenges when studying gestures is to understand their interactional context and to classify them as relevant or irrelevant gestures to a specific interactional context in order to take a corresponding interactional reaction.

### (1) Classes of gestures

Gestures could be classified in terms of the interactional context to [37] [38] [40] [41]:

### (1-1) Irrelevant gestures

These gestures include irrelevant motion and side effects of the human motor behavior. They are neither communicative nor socially interactive. They may be salient, but are not movements that are primarily employed to communicate or engage a partner in interaction, like: motion of the arms and hands when walking, tapping of the fingers, playing with a paper clip, brushing hair away from the face with the hand …etc.

### (1-2) Side effect of expressive behaviour

These gestures are used spontaneously when communicating with others including: the motion of hands, arms and the facial expressions in order to reinforce the meaning of the interactional message.

### (1-3) Symbolic gestures

These gestures could be considered as conventionalized signals in a communicative interaction. They belong generally to a limited, circumscribed set of gestural motions that have specific and prescribed interpretations. They are used to trigger certain actions according to a code or convention, like: waving a taxi to stop, nodding *Yes* or *No*, waving a greeting *Hello* or *Goodbye* …etc.

An important issue to take into consideration is that the interpretation of some conventionalized gestures may differ from a place to a place or from a country to a country according to the domestic culture of that place or country. Therefore, one of the main problems facing the gestural interaction between a robot and a human on a wide range, is the difficulty to generalize a total conventionalized set of gestures regardless of the place or the culture, example: the gesture *OK* indicated in the following figure means in USA or England that every thing goes well, while in Latin America, it means an insult!!

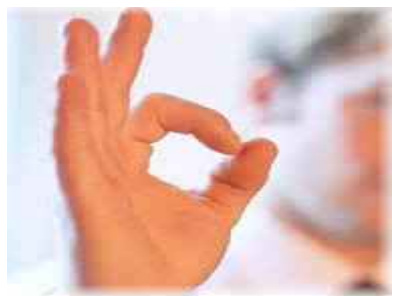

**Fig. (A2-1)** Gesture OK

Similarly, the meaning or the characteristics *(speed, frequency .etc)* of the gestures could be different regionally even within the same country. So, it is clear how huge is the difficulty of making an autonomous robot able to perfectly interact with humans regardless of the domestic culture of the place, region or country.

### (1-4) Interactional gestures

These gestures are used to regulate the interaction with a partner, i.e. used to initiate, maintain, invite, synchronize, organize or terminate a particular interactive, cooperative behaviour, like: raising an empty hand towards a partner inviting him to put something in it, raising a hand containing an object towards a partner inviting him to take it, nodding the head to a partner to confirm listening to him etc.

The important point in this type of gestures is that they don't convey any content of the communication *(like in the Side Effect of Expressive Behaviour gestures)*, but instead, they are used to regulate just the interaction between humans or human-robot as indicated above.

### (1-5) Pointing gestures

These gestures are used to refer to or to indicate objects of interest in a communication using hands, fingers, and/or eye gaze.

## (2) Gestures recognition in the human robot interaction

In the context of the human robot interaction, gestures' classification is indispensable in order to allow the robot to make the suitable reaction. Gestures belong to the above 5 classes seem to be of different levels of difficulty in classification. Symbolic and pointing gestures seem to be the most easy to be classified because they are relatively limited with respect to the other classes and they have precise forms easy to be characterized and classified on a wide range.

The interactional gestures' class seems to be bigger than the symbolic and pointing gestures' classes, but these gestures still have specific forms; consequently, they could be well classified especially if the interactional context is precised.

Last but not least, irrelevant and side effect gestures seem to be the most difficult gestures to characterize and to classify because their classes are highly unlimited and they are implemented in a random sequence according to the person who performs them. Therefore, there is no possibility to train a recognition system on such random and unlimited classes or else the database has to be enormously big which will affect badly the processing time and so the interaction.

The interactional history may be important and useful in such case to determine the possible class of the gesture, also, it could be useful to precise the interactional scenario between the human and the robot because it is not possible *for the moment* to have an autonomous robot that can interact in an absolute way with humans regardless of the type of the gesture or the context of interaction.

## (3) Special difficulties during gestures recognition

An important problem may takes place during gestures' recognition, is when a gesture could be classified in two classes, which is known as the *intermediate* gestures, like; holding a yellow card, which could be classified to the irrelevant gestures or the symbolic gestures. In order to resolve this problem, it is required to take into consideration the environment and the context of the gesture. So, if the interactional context of the gesture is, for example, a football game, the gesture will be considered as a symbolic gesture, otherwise, it will be considered as an irrelevant gesture.

In my practical work, i constructed a database composed partially (*) of pointing and symbolic gestures to analyze. In the future, this database could be generalized to include other classes of gestures like the interactional gestures, according to the advice of the medical team who will choose certain gestures to teach to autistic children using social robots through the proposed triadic *(Human – Robot – Child)* imitational games. Irrelevant and side effect gestures, in my opinion, will never be included to the database, because the goal of using social robots in the case of autism, is to interact with children in a supervised way via a prescribed scenario not in an absolute or random way, and so, irrelevant and side effect gestures are meaningless in this context, cause they don't have any usefulness in ameliorating the social interactional behavior of these children.

---

(*) There are other gestures in my database which don't belong actually to any of the above studied 5 classes, like walking from left to right and vise versa, and moving the right arm to down from up position. I added them to my database just to test and prove the relevance of the characteristic vector, and to see how will be the classification results. These gestures will not be used in any interactional scenarios with autistic children because there is no use of them in this context as long as they don't convey any precise communicating messages.

# Appendix 3

## Autism

Autism, first described by American psychiatrist *Leo Kanner* in 1943, is thought to result from a brain disorder that takes place during the first two and a half years of childhood. It is characterized by social disconnectedness, failure to recognize and read the subtleties of human communication behaviors and interactions, an obsessive addiction to routines and repeatable behaviors, and the repetition of sentences and words without regard to their significance or the context in which they are spoken.

## (1) Identification

As there is no clear biological marker of autism, it is always not easy to identify it. The identification depends only on the observation of the psychiatrist to the child basing on the characteristics of autism mentioned above [42] [43] [44].

### (1-1) Factors to take into consideration while identifying autism

- The manifestation of autism can change with age, as patterns of atypical behavior and impairments become more or less marked over time.
- Actual capacity in any one individual may fluctuate at different times and in different circumstances.
- The severity of autistic characteristics varies widely across the autism spectrum.

| Behavioral variation could be varied | | |
|---|---|---|
| | **From** | **To** |
| **Key characteristics** | | |
| Social interaction | Aloof and indifferent | Makes one-sided approaches |
| Social communication | No communication | Spontaneous but repetitive, one-sided |
| Repetitive behavior/activities | Simple, bodily directed (e.g. face tapping, self injury) | Verbal, abstract (e.g. repetitive questioning) |
| **Other characteristics** | | |
| Formal language | No language | Grammatical but long winded, repetitive, literal interpretations |
| Responses to sensory information (oversensitivity, insensitivity) | Very marked | Minimal or no unusual responses |
| Unusual movements (hand flapping, tiptoe walking) | Very marked | Minimal or absent |
| Special skills (drawing, rote memory) | None | One skill at a high level, very different from other abilities |

**Table. (A3-1)** Variation in behavior across autism spectrum

From the table above we can notice that some autistic people have remarkable talent in a specific field; like that autistic child who is able to make the most complicated mathematical calculations with his mind in few seconds, etc. But, despite that fact that autistic people may be talented in some specific fields, they have other normal or weak responses concerning other interactive abilities. So, the importance of using social robots comes clear to mind, hoping that the weak points in the interactional behavior of an autistic child or adult get developed, but at the end autistic people will stay autistic at least up till now, may be in the future if the real causes of autism get discovered that may lead to a medical or psychological treatment for it.

## (2) Possible causes of autism

Some reasons are thought to be causes of autism, like [42] [46]:

### (2-1) Genetics

Some studies searched the relationship between autism and the genetic kind as a male or female, taken together with studies looking at families and twins. The results confirmed that autism has a genetic component. However, the exact mechanism by which genes are implicated in autism is unclear and is an important focus for future research. In addition, future studies aim to determine how genes interact with environmental factors in autism.

### (2-2) Biology of the brain

Progress in identifying brain differences between the brains of autistic and non-autistic people has been slow and findings are inconsistent. Nevertheless there is evidence suggesting that in autism:

- Brains are larger and heavier and there are differences in the cells of some brain regions.
- There is reduced activity in areas associated with the processing of social and emotional information,
   planning and control of behavior.
- There are differences to some signaling molecules in the brain such as serotonin.

As a result of the inconsistent biology of the brain, autistic people have difficulties in planning, controlling behavior and understanding the mental states of others.

## (3) Interventions

It is generally agreed that no single intervention will suit all autistic people, and that any intervention can have negative as well as positive effects. A range of interventions have been developed. Examples include those based on behavioral methods, education-based approaches and non-verbal communication systems. Pharmacological interventions have also been used with some success to treat depression and anxiety in autism, although they do not treat autism itself, and till now there is insufficient evidence about which autism interventions will be effective and why, which reflects the current situation of medical advance towards autism and how it is a complicated issue.

## (4) Priorities of research

Nowadays, the autism research funded projects are focusing mostly on the symptoms (including brain differences, mental processes and behavior, about 53 % of the whole funding) more than epidemiology (4 % of the whole funding) and causes (17 % of the whole funding), as indicated in the figure below.

This simple statistic reveals that the medical advance with respect to autism still retarded to the extent that makes the researching budget be focused on discovering more and more the symptoms of autism to define and categorize them in a more precise way.

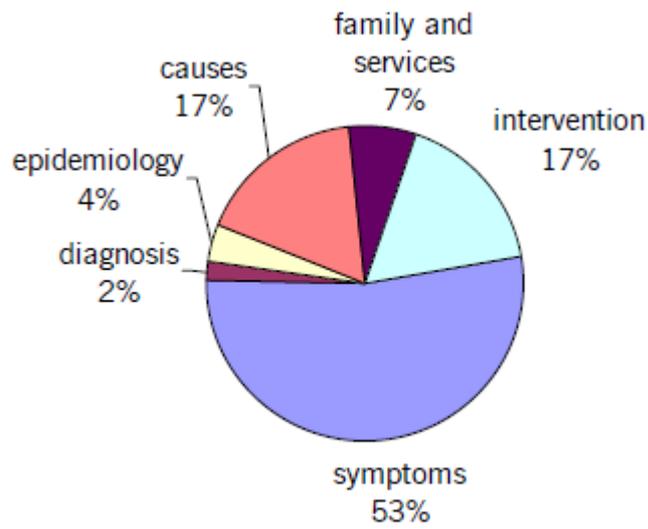

**Fig. (A3-1)** Breakdown of autism research funding

    Last but not least, the autistic child or adult will need a special care and services throughout his life. These supporting services vary according to the age (childhood or adulthood) and the medical situation of the autistic child or adult ranging from ability to disability, so, they always needs an attentive supervision to help them communicate positively with the external environment .

# Appendix 4

## Human robot interaction experimental design

- *This part was developed in close collaboration with Prof. Adriana Tapus.*

The consideration of robots as social tools is still a new area of scientific research. Intelligent assistive robotics is a new area of the Human Robot Interaction *HRI* that tries to bring together a broad spectrum of research including robotics, medicine, social and cognitive sciences, and neuroscience among others.

The following figure illustrates an interactive architecture between the human and robot. This architecture presents the signal flow between the human and robot as well as the reasoning and decision making algorithms to manage an interaction. We aim at individualize one-on-one social interaction and triadic interaction (through joint attention, games, imitation) by using multimodal communication (speech, gestures, body movements). To ground the proposed research in a real-world domain, we include validation with human subjects from both typical populations and those with Autism Spectrum Disorders *ASD*.

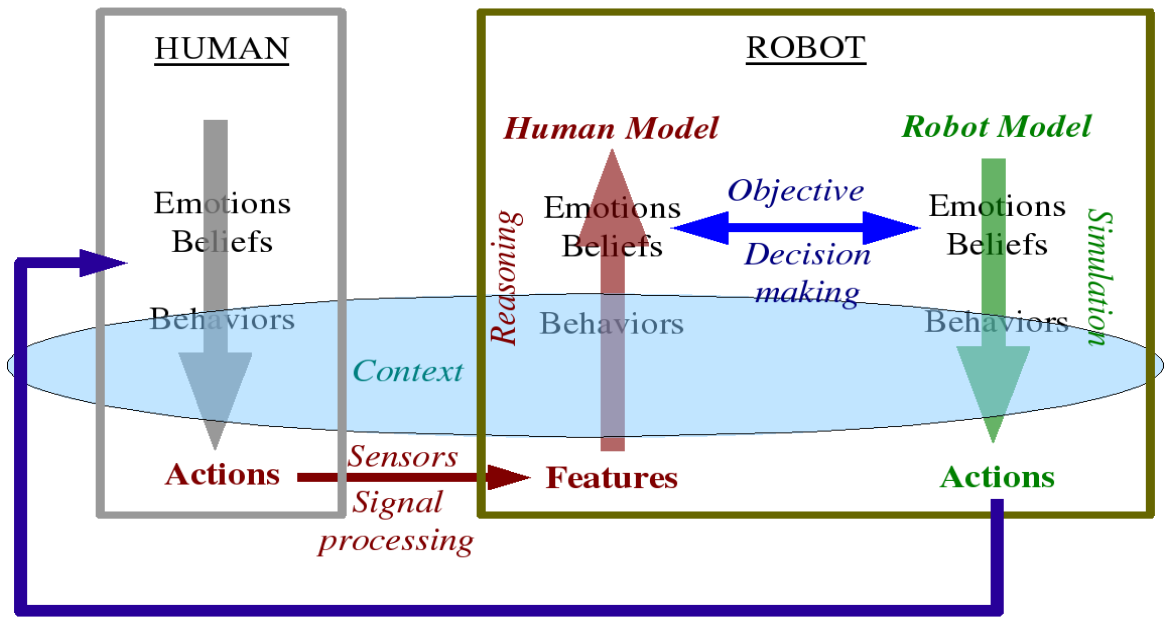

**Fig. (A4-1)** *HRI* architecture

## (1) Protocol of experiments *(Case study)*

This part focuses on an imitating scenario to perform and validate with autistic children under the supervision of a medical team and the cognitive robotics team in *ENSTA Paris-tech*. The following scenario is based on engaging the autistic child to an attractive game which is proved to be the best way to make the child interact positively with the robot.

### (1-1) Test Bed

Robot *NAO*, developed by Aldebaran robotics, is supported by different abilities to communicate with external computers to analyze the video and audio data captured by the sensors *(see appendix 1)*. These external computers will, in turn, control the robot in order to perform a correspondent action to the analysis done.

## (1-2) Design *"Robot's help in motor imitation for children with autism"*

In this study, we try to observe:

- Whether the robot can induce imitation in the user, and sustain an imitation game over time. Mimicry and imitation are powerful tools of human engagement; we will test them in the *HRI* context.

Motor imitation problems are common in autism. Imitation is a combination of cognitive representational and visual perceptual-motor components. Motor imitation can be divided into meaningful transitive and intransitive gestures versus non-meaningful single and sequential hand postures. Meaningful gestures are symbolic and language related. In this study we focus only on intransitive gestures, e.g. wave goodbye, which are communicative gestures. Imitation of meaningful gestures requires a proper visual perceptual discrimination of the action, comprehension of the meaning of the action, and linkage to previous information about the 'time–space' formulas of a to-be-performed action. The imitation of meaningful gestures requires cognitive-representational as well as perceptual-motor processes [Rogers and Pennington, 1991 [49]; Vanvuchelen et al., 2007 [50]].

Teaching imitation, and demonstrating the behavior of an autistic child through imitation, have both led to improved social responsiveness. Use of language and gestures within the zone of proximal development *(i.e., made possible in collaboration with others)* may encourage the child to use the same form themselves.

Moreover, it is well-known that the task performance of children with autism (known for their low task motivation – [Koegel and Mentis, 1985 [51]]) enhanced when their task involvement was increased. It is important to mention that the performance of children improved, for example when they were prompted by explicit instructions, when their active task participation was reinforced, or when their personal interest increased [Begeer, 2003 [52]].

The purpose of this study is to use the triadic interaction between child, robot and therapist/family member so as to help and improve children's performance on the imitation on meaningful gestures and to be more precise on intransitive gestures which are known as communication gestures *(see Figure 2)*.

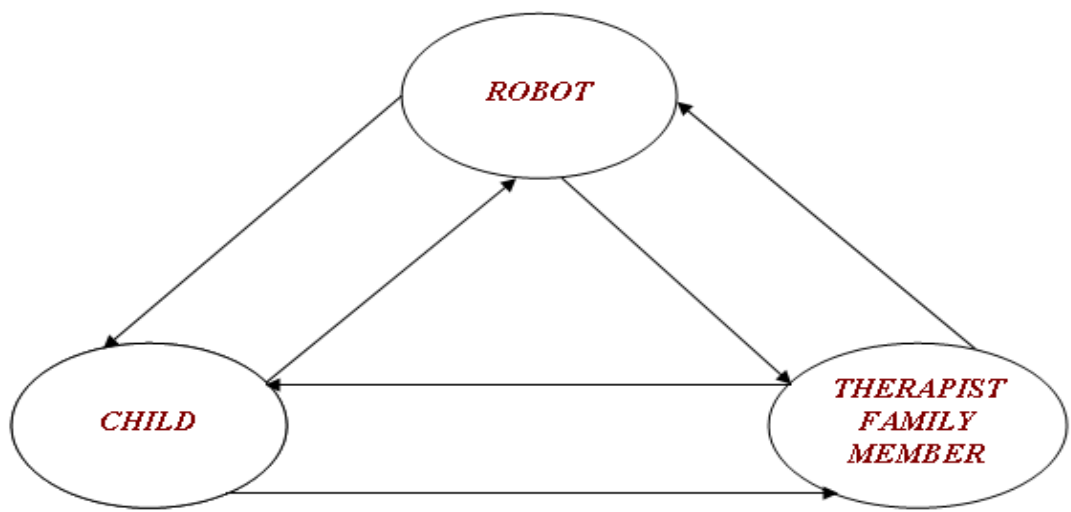

**Fig. (A4-2)** Triadic interaction

In this study, the robot will serve as a social conduit to facilitate and mediate social interaction with other, unfamiliar and familiar people.

### (1-2-1) Hypotheses

1. The robot from the triadic interaction can improve the task performance of the child's imitation skill with respect to the dyadic interaction.

2. During the course of the triadic interaction, the child can improve his/her verbalisation.

### (1-3) Participants

We aim at having two groups of children with *ASD*: One involved in the triadic interaction and the other involved only in the child-therapist interaction.

### (1-4) Scenario

We will validate our approach and implementation on a robot system designed for use with both typical children and *ASD* children. The experiments will be designed around the following scenario, designed to stimulate specific social behaviors through *HRI*.

### (1-4-1) Orientation session

Before starting the experiments we will have a first session that corresponds to an introductory *(orientation)* phase that will permit to children to become familiar to the *NAO* robot and the different robot's movements. In this first session, the participant will be 'introduced' to the robot. The *NAO* humanoid robot will be brought into the room with the participant, but will not be turned on. During this introduction period, the family member/therapist will explain the robot behavior, the overall goals and plans for the study, and generally inform the participant of what to expect in future sessions.

Finally, at the end of the first session, the therapist will provide or will run cognitive tests with children so as to determine the level of cognitive impairment *(related to joint attention, verbalization, motor imitation, emotions understanding, and their capacity to socially relate and interact to people)*.

The primary purpose of this first experiment is to determine the participant's initial, no robot condition motor task functionality, mental state, and level of cognitive impairment. All the experiments will be video and audio recorded. The total duration of this first session is of maximum 15 minutes *(if no cognitive tests are run)*.

### (1-4-2) Imitation Game

Children are instructed as follows:

"We will play a game. The three of us *(you, the robot and the therapist/family member)* will sit in a circle. Each of us will make a gesture and the other person to the left *(or to the right)* will have to imitate the same gesture. So for example, the robot or the therapist will move once the hand and you will have to reproduce the same movement. Once this is done you will make a gesture and the robot or the therapist will imitate you."

The robot or the therapist will also:

1. Prompt with explicit instructions when the child turn would come.

2. Congratulate for the gesture initiated/produced.

This game gives control to the child through which he will be able to control the robot's/therapist's movements with his own. Imitation will be used as means of capturing and holding child's attention. The robot will map the observed behavior of the user to its own behavior repertoire and perform the best possible imitation. The ability to imitate will be used to induce the user to mimic the robot in return.

The time on task/game can be a fixed amount of time *(10-15 minutes)* or can be open ended. Scoring criteria can be based on a pass or fail system for each gesture reproduced:

1. With *0* reflecting no action at all or an action that appeared unrelated to the target action.

2. With *1* reflecting an action related to the target action modelled by the robot/therapist.

### (1-5) Data Collection

Multi-modal data will be collected from the trials. The interaction will be video recorded. The cameras *(both the ones on the robot and the ones in the room)* will record both the child's face, movements, and interactions, and will also record the robot and the third person in the scene *(for the triadic interaction)*. A microphone will also be used to record children's verbalization. Videos and speech data will be annotated for more analysis.

# Appendix 5

## K-nearest neighbor algorithm

The k-nearest neighbor *(KNN)* rule is one of the oldest and simplest classification rules: for a given query vector $x_o$ and a set of *N* labeled instances $\{x_i, y_i\}_1^N$, the task of the classifier is to predict the class label of $x_o$ on the predefined *P* classes. The k-nearest neighbor *(KNN)* classification algorithm tries to find the k nearest neighbors of $x_o$ and uses a majority vote to determine the class label of $x_o$.

Without prior knowledge, the *KNN* classifier usually applies Euclidean distance as the distance metric. However, this simple and easy-to-implement method can still yield competitive results even compared to the most sophisticated machine learning methods.

The performance of a *KNN* classifier is primarily determined by the choice of k as well as the distance metric applied. However, when the points are not uniformly distributed, predetermining the value of k becomes difficult.

The coming figure, which depicts a 3-nearest neighbors' classifier on a two class problem in a two dimensional feature space, indicates more the idea of the *KNN* classification. In this figure, the decision for $q_1$ is straightforward; all its 3 neighbors are of class *O* so it is classified as an *O*. While, the situation for $q_2$ is a bit more complicated as it has two neighbors of class *X* and one of class *O*, this can be resolved by simple majority voting or by distance weighted voting, i.e. $q_2$ will be attributed to class *X*.

So, *KNN* classification has two stages; the first is the determination of the nearest neighbors and the second is the determination of the class using those neighbors [53].

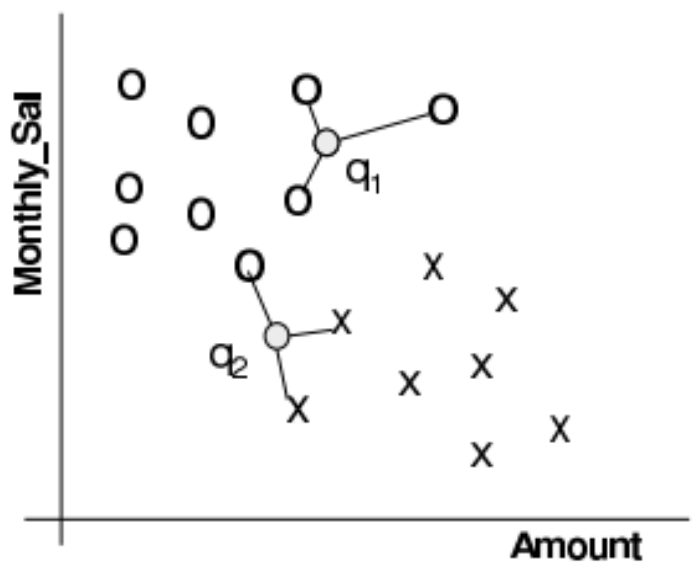

**Fig. (A5-1)** A simple example of 3-Nearest Neighbors classification

Let us assume that we have a training dataset $D$ made up of $(x_i)_i \in [1, |D|]$ training samples. The examples are described by a set of features $F$ and any numeric features have been normalized to the range [0, 1]. Each training example is labeled with a class label $y_i \in Y$. Our objective is to classify an unknown example $q$. For each $x_i \in D$ we can calculate the distance between $q$ and $x_i$ as follows:

$$d(q, x_i) = \sum_{f \in F} w_f \, \delta(q_f, x_{if}) \qquad (1)$$

There are a large range of possibilities for this distance metric; a basic version for continuous and discrete attributes will be:

$$\delta(q_f, x_{if}) = \begin{cases} 0 & f \text{ discrete and } q_f = x_{if} \\ 1 & f \text{ discrete and } q_f \neq x_{if} \\ |q_f - x_{if}| & f \text{ continuous} \end{cases} \qquad (2)$$

The k-nearest neighbors are selected based on this distance metric; then, there are a variety of ways in which the k-nearest neighbors can be used to determine the class of $q$. The most straight forward approach is to assign the majority class among the nearest neighbors to the query. It will often make sense to assign more weight to the nearer neighbors in deciding the class of the query. A fairly general technique to achieve this is distance weighted voting where the neighbors get to vote on the class of the query case with votes weighted by the inverse of their distance to the query.

$$Vote(y_j) = \sum_{c=1}^{k} \frac{1}{d(q, x_c)^n} 1(y_j, y_c) \qquad (3)$$

Thus the vote assigned to class $y_j$ by neighbor $x_c$ is 1 divided by the distance to that neighbor, which returns 1 if the class labels match and 0 otherwise. In equation *(3)*, $n$ would normally be 1 but values greater than 1 can be used to further reduce the influence of more distant neighbors.

# Appendix 6

## Cross validation

It is a technique for assessing how the results of a statistical analysis will generalize to an independent data set. One round of the cross validation involves partitioning a sample of data into complementary subsets, performing the analysis on one subset *(the training set)*, and then, validating the analysis on the other subset *(the testing set)*. To reduce variability, multiple rounds of cross-validation are performed using different partitions, and the validation results are averaged over the rounds. The basic form of cross-validation is the k-fold cross-validation. Other forms of cross-validation are special cases of the k-fold cross-validation or involve repeated rounds of the k-fold cross-validation. Generally, the k-fold cross-validation is the used cross validation method in this work, and will be discussed clearly in this appendix.

## (1) K-fold cross validation

In k-fold cross-validation the data is first partitioned into k equally (or nearly equally) sized segments or folds. Subsequently, k iterations of training and validation are performed such that within each iteration, a different fold of the data is held-out for validation while the remaining k-1 folds are used for learning and so on.

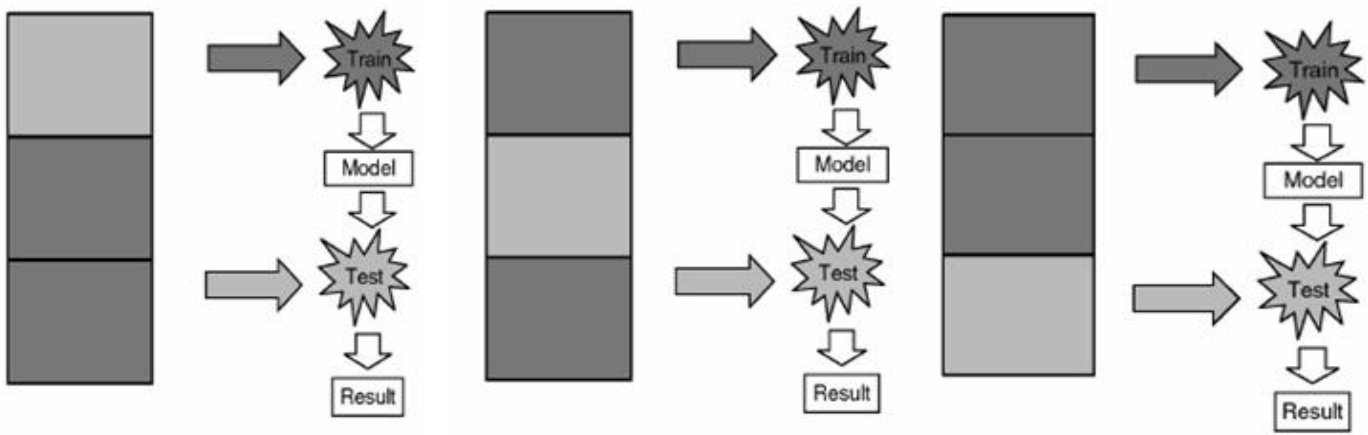

**Fig. (A6-1)** Procedures of three fold cross-validation

The previous figure demonstrates an example with k = 3. The darker section of the data is used for training while the lighter sections are used for validation. In data mining and machine learning 10-fold cross-validation (k = 10) is the most common [54].

## (2) Key applications of cross validation

Cross-validation can be applied in the following contexts: performance estimation, and model selection

### (2-1) Performance estimation

As previously mentioned, cross-validation can be used to estimate the performance of a learning algorithm in terms of accuracy and precision. It allows for all the data to be used in obtaining an estimate for the recognition score of a learning algorithm.

Using *(for example)* 10-fold cross-validation, there will be a bout 90% of the data to build a model and learn it, while 10% of the data will be used for testing the accuracy. The resulting average accuracy is somewhat representative for the true accuracy when the model is trained on all data and tested on unseen data, but in most cases this estimate is reliable, particularly if the amount of labeled data is sufficiently large and if the unseen data follows the same distribution as the labeled examples.

**(2-2) Model selection**

Cross-validation may be used to compare a pair of learning algorithms. This may be done in the case of newly developed learning algorithms, in which case the designer may wish to compare the performance of the classifier with some existing baseline classifier on some benchmark dataset.

Therefore, the advantage of the cross validation seems so clear in the context of validating the recognition and classification process, and also, in order to obtain a general constant recognition score for the entire database. Consequently, it was used side to a side to the k-nearest neighbor classification method, which i used frequently in this work.

# Appendix 7

## Controlling the robot NAO

The controlling process of the robot *NAO* has two sides; the side of the external pc and the side of the robot. These two sides are communicating between each other via *TCP* socket communication, as indicated in the coming figure.

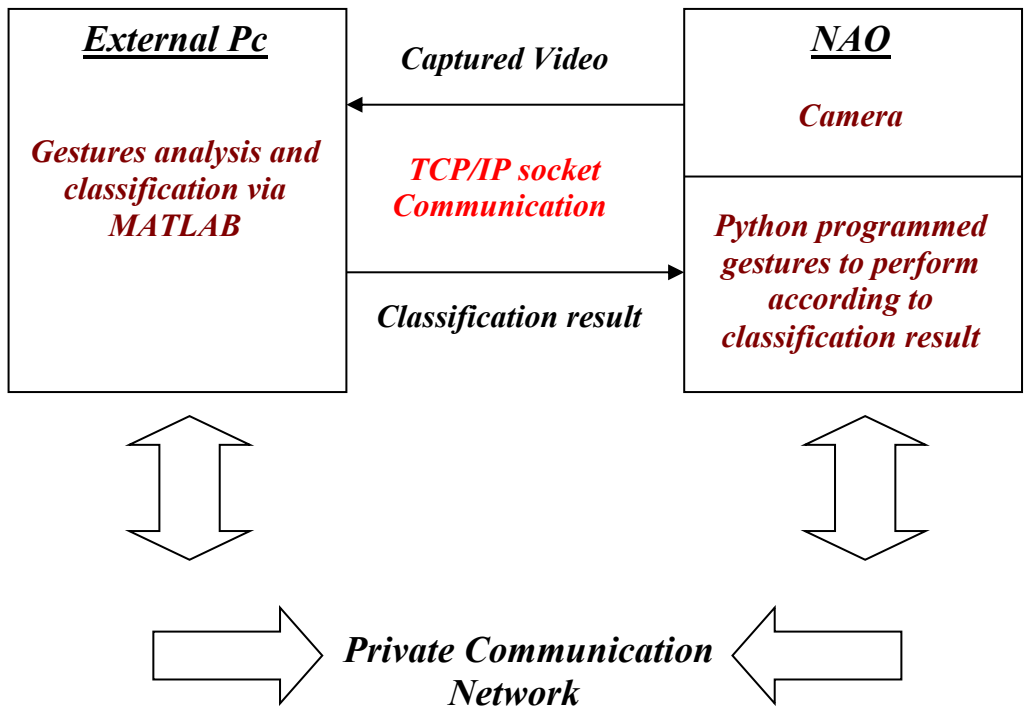

**Fig. (A7-1)** Outline for the PC/ Robot Interface

The previous outline reveals that the robot camera will capture a video according to a prescribed number of frames, and then sends it via *Client/Server TCP/IP* socket to the external computer working on *MATLAB*. *MATLAB* automatically start analysing the captured video and send back the classification result via *Client/Server TCP/IP* socket to the robot side, which has some pre-programmed gestures controlled by python programming language according to the coming classification result.

The communication takes place over a private communication network where the robot and the computer takes suitable *IP* addresses which enables the socket communication to start moving data in both directions with no problems.

Actually the private communication network can transfer data via a communication speed between 8 to 12 Mega bytes / second. So the process of sending a video of an average size of 1 Mega bytes or so, takes few parts of a second. The analysis of the captured test video depends on the state of the PC which has to be of good processing abilities to take about 2 seconds for giving the classification result. That is to say that, the total time from the moment of capturing the video to the moment of imitating the gesture will not exceed 3 or 4 seconds maximum which is a good time response.

During the research I did in the laboratory, and in order to test the recognition and imitation practically using the robot, I choosed some testing gestures made by a different person with respect to other gestures in the database, then i filmed these gestures by an external camera and sent these gestures to my PC to be tested (*by the time, i haven't use the camera of the robot yet)*. The number of the choosed testing gestures was 6, therefore, in order to control the robot, i constructed a vector of six components and they were all equal to zero. When the program reveals the class of the tested gesture, its correspondent component in the vector gets equal to one and the other components rest equal to zero, ex; suppose the class of the tested gesture =3 , therefore the controlling vector will be written as following : *Control vector = [0,0,1,0,0,0]*.

Afterwards, it changes the vector to a string and sends it to the robot via the *TCP* socket. From the robot side, the programmed server on python language will receive that string and will change it again from string to vector and then, will search the component of the vector which equals to one. When it finds it, it gets its order in the vector (*order =3)*, and performs the third preprogrammed gesture of the whole programmed group of gestures on the robot. The results obtained of the recognition and imitation in this small test was 100%; i.e. the robot understood well the gesture and performed it perfectly and that was the main practical target of the whole work.

# REFRENCES